%% file: main.tex
\documentclass[10pt,journal,compsoc]{IEEEtran}
\usepackage{amsmath,amsfonts}
\usepackage{algorithmic}
\usepackage{algorithm}
\usepackage{array}
\usepackage[caption=false,font=normalsize,labelfont=sf,textfont=sf]{subfig}
\usepackage{textcomp}
\usepackage{stfloats}
\usepackage{url}
\usepackage{verbatim}
\usepackage{graphicx}
\usepackage{cite}
\usepackage{ig}
\labelfigure{IG}
\usepackage{tikz}
\usetikzlibrary{calc,arrows,shapes}
\def\Tom#1{{\color{red}#1}}
\def\Tom#1{#1}
\usepackage{hyperref}
\hypersetup{
    colorlinks=true,
    linkcolor=blue,
    filecolor=blue,      
    urlcolor=blue,
    citecolor=blue,
    pdfpagemode=FullScreen,
    }
    
\usepackage{acronym}
\acrodef{NLP}{natural language processing} 
\acrodef{AI}{artificial intelligence} 
\acrodef{ML}{machine learning} 
\acrodef{CNN}{convo\-lu\-tional neural network} 
\acrodef{RNN}{recurrent neural network} 
\acrodef{LSTM}{long-short term memory} 
\acrodef{LLM}{large language model} 
\acrodef{DWC}{dictionary word count} 
\acrodef{IG}{integrated gradient} 
\acrodef{PoS}{part of speech} 
\acrodef{TF-IDF}{term frequency-inverse document frequency}

\begin{document}

\title{Application of integrated gradients explainability to sociopsychological semantic markers}

\author{Ali Aghababaei, Jan Nikadon, Magdalena Formanowicz,\\ Maria Laura Bettinsoli, Carmen Cervone, Caterina Suitner, Tomaso Erseghe 
\thanks{A. Aghababaei and T. Erseghe are with the Department of Information Engineering, University of Padova, Padova, Italy, Contact: tomaso.erseghe@unipd.it}
\thanks{J. Nikadon and M. Formanowicz are with the Center for Research on Social Relations, University of Social Sciences and Humanities (SWPS), Warsaw, Poland. J. Nikadon and M. Formanowicz contribution was financed by the OPUS 19 grant of the Polish National Science Center (2020/37/B/HS6/02587)}%
\thanks{J. Nikadon  is with the Department of Cognitive Science, Nicolaus Copernicus University in Toruń, Poland and Interdisciplinary Centre for Modern Technologies, Nicolaus Copernicus University in Toruń, Poland}%
\thanks{M.L. Bettinsoli, C. Cervone and C. Suitner are with the Department of Developmental Psychology and Socialization, University of Padova, Padova, Italy}
\thanks{This work has been submitted to the IEEE for possible publication. Copyright may be transferred without notice, after which this version may no longer be accessible.
}}

\markboth{Submitted to IEEE Transactions on Computational Social Systems}%
{Submitted to IEEE Transactions on Computational Social Systems}

\IEEEpubid{0000--0000/00\$00.00~\copyright~2026 IEEE}

\IEEEtitleabstractindextext{%
\begin{abstract}
Classification of textual data in terms of sentiment, or more nuanced sociopsychological markers, 
is now a popular approach commonly applied at the sentence level. In this paper, we exploit \Tom{reliable attribution methods} to capture the classification output at the word level, revealing which words actually contribute to the classification process. This approach improves explainability and provides in-depth insights into the text. We focus on sociopsychological markers beyond sentiment and investigate\Tom{, as a test case,} how to effectively train algorithms in agency, one of the very few markers for which a validated deep learning classifier, BERTAgent, is currently available. The performance \Tom{of different solutions is carefully tested, algorithm parameters are optimized, and the \ac{IG} method is finally selected for both enhanced precision and controlled computational complexity. The reliability} of the result is \Tom{validated, and then applied to} relevant application scenarios, including one where a small labeled dataset is available, with the aim of exploiting IG to identify the salient words that contribute to building the different classes that relate to relevant sociopsychological markers. To achieve this, a training procedure that encourages overfitting is employed to enhance the distinctiveness of each class. The results are analyzed through the lens of social psychology.
\end{abstract}

}

\maketitle

\section{Introduction}

\IEEEPARstart{E}{xtracting} semantic information, for example, sentiment, from textual data is now a widespread application of \ac{NLP} techniques based on available deep learning tools \Tom{and \acp{LLM}}. It is widely applied in various fields, including advertising, financial risk assessment, consumer insights, and government affairs, among others \cite{poria2023sentiment}. The literature on these topics is vast, including, only for the keyword ``sentiment classification,'' more than 2,000 articles per year in the last five years (source: Scopus). 
\IEEEpubidadjcol

The deep learning models in use span from \acp{CNN} to \acp{RNN} and \acp{LSTM} \cite{seo2020comparative}. However, the most powerful techniques \Tom{are \acp{LLM}} exploiting the so-called ``Transformer architecture,'' using an attention-based deep learning mechanism \cite{vaswani2017attention}  where the language model (e.g., BERT \cite{devlin2018bert}, RoBERTa \cite{liu2019roberta}, or GPT \cite{radford2018improving, radford2019language, brown2020language}) is pre-trained on huge datasets and is therefore able to fully interpret the complexity of language. The undisputed superiority of pre-trained algorithms based on the Transformer architecture is evident from the literature \cite{alaparthi2021bert}.\footnote{For sentiment-like detection and classification there exist many alternative methods, other than deep learning, that can be used, such as the widely used and validated \ac{DWC} approach of \cite{pennebaker2001linguistic}. However, these methods lack the power of large language models (e.g., contextual information at the sentence level, automatic management of negations, etc. \cite{nikadon2023}). For this reason, they are not always effective and, therefore, are not considered.} Such models have been used to identify sentiment, but also other linguistic features, for example emotions such as anger, sadness, and happiness \cite{peng2022survey,deng2021emotion}, or more nuanced aspects such as offensive language, hate speech, sarcasm, and irony \cite{abdullah2022deep,ataei2023pars}. More recently, these models have been applied to the identification of agency, which is the focus on goal orientation and achievement \cite{nikadon2023}. 
However, numerous other theoretical constructs such as sexualization, affection, depression, and affiliation can be associated with linguistic markers (see also \cite{pennebaker2001linguistic}).

Available approaches commonly evaluate sentiment, or any other linguistic marker, at the sentence or document level, meaning that each text is classified into a single value. Although extremely valuable, this approach has a limitation: we are not aware of what part of the text (i.e., which words) effectively contributes to the prediction, and the role of text elements (i.e., words) is obscured by the complex deep learning mechanism. However, this information could be highly valuable as an additional signal for researchers, as it provides a more readable result in an ``explainable AI (XAI)'' manner. Using XAI methods helps to determine how specific terms influence the sentiment of a text. These methods offer several benefits, including increasing model accountability for their decisions, providing insight into factors that shape sentiment, and supporting industries such as finance and healthcare, where regulatory frameworks require explanations of model outputs \cite{SentimentXAI2024}. 

A solution to this problem  \Tom{can be obtained through a number of state-of-the-art methods available from the literature. Among these,} the \ac{IG} approach is widely used for interpretable AI recommendations due to its interesting properties: it measures the importance of features by averaging the model output gradient along a straight path and is computationally efficient. Furthermore, it has the desired completeness property, which means that the attributions add up to the target output when a baseline is carefully chosen \cite{linardatos2020explainable}. Although \ac{IG} is classically applied to images, recent research has highlighted its role in \ac{NLP}, especially in relation to sentiment analysis \cite{bartivcka2022evaluating, santoso2023text, grosz2023discovering, zielinski2023dataset}. Improvements to the baseline \ac{IG} approach are also available, for example, the sequential-\ac{IG} approach, which computes the importance of each word in a sentence by keeping all other words fixed and only generating interpolations between the baseline and the word of interest \cite{enguehard2023sequential}, or the discretized-\ac{IG} approach, which uses a more effective attribution along non-linear interpolation paths \cite{sanyal2021discretized}. Methods alternative to \ac{IG} exist for interpreting the Transformer architecture, such as DeepLIFT \cite{shrikumar2017learning} or SHAP \cite{mosca2022shap, scott2017unified}.

In this paper, \Tom{we analyze and compare the effectiveness of state-of-the-art solutions, being thus able to identify \ac{IG} \cite{sundararajan2017axiomatic} as the most effective approach in terms of both performance and complexity. We specifically exploit it} to access entity-level information (i.e., word-level \Tom{attributions}) for sociopsychological markers other than sentiment, and focus on agency, an important construct, one of the very few markers for which a pre-trained and validated \ac{NLP} predictor, BERTAgent \cite{nikadon2023}, is available. We work on BERTAgent to both optimize the \ac{IG} parameters and to verify the suitability of the \ac{IG} approach for the task. 

This part covers a scenario where large labeled datasets are available for adequate training of the chosen sociopsychological marker (e.g., agency in the BERTAgent case). However, since a question arises as to what can be done in the most common situation where only labeled datasets of limited or very limited size are accessible (as large labeled datasets are not always available and are often difficult to obtain), we further demonstrate the usefulness of the \ac{IG} approach in these cases for a diverse collection of alternative markers. Specifically, we devise an \ac{NLP} training procedure aimed at classification that, unlike the common approach, \emph{encourages overfitting} so that the subsequent application of the \ac{IG} rationale can identify the words that best align with the identification of the different classes (e.g., different levels of markers such as positive, negative, and neutral). In this way, we show how \ac{IG} is able to provide a detailed description of the classes and, more importantly, of the salient words that differentiate the classes. This is a valuable asset for data explainability and can be particularly useful for building those dictionaries that constitute the starting point in the creation of large labeled datasets. 

The remainder of the paper is organized as follows. \Tom{In \sect{AG} we introduce our case study for socio-psychological investigation of text, that is agency and its meaning, as well as the verified \ac{LLM} tool BERTAgent. In \sect{MM} we discuss all the attribution methods that will be compared in the paper, as well as the datasets used. \sect{IG} presents the performance comparison among different attribution methods, as well as the reasons for selecting \ac{IG}. The reference context is one} where a large dataset is available to build a reliable \ac{NLP} model. For this task, we selected the pre-trained BERTAgent agency predictor available from \cite{nikadon2023} and based on RoBERTa.\footnote{We incidentally note that recent studies actually highlighted the effectiveness of RoBERTa over other PLMs in sentiment analysis and emotion detection \cite{mao2023biases}.} We use it as a test case to identify the most relevant system parameters for \ac{IG}, namely, integration step and baseline, as well as to compare the performance of \ac{IG} with state-of-the-art alternatives using sufficiency, compactness, and comprehensiveness criteria \cite{wang2022fine}. \Tom{The reliability of the chosen \ac{IG} approach is validated in \sect{VE}. The concluding sections present some application examples. In \sect{AO}} highlights in text prepared by experts are compared to the outcome of the \Tom{BERTAgent-based} \ac{IG} approach to assess the extent to which the highlighted sections, which pertain to collective action intentions and heavily rely on agentic motives \cite{van2016building}, are filled with agentic content. The application of \ac{IG} under a small dataset scenario is presented in \sect{TO}, where alternative sociopsychological markers are discussed, to demonstrate the potential of the proposed approach \Tom{for building or extending dictionaries on those classification tasks lacking a strong and mature theoretical ground (e.g., objectification)}. An interpretation through the lens of sociopsychological theories further emphasizes the usefulness of the method. \sect{CO} concludes the paper.

\Section[AG]{Case study: Agency and BERTAgent}

In this paper we mainly concentrate on a specific sociopsychological marker called agency. Agency is defined as the capacity for goal orientation and the ability to plan and execute goal achievement \cite{bakan1966duality, bandura2001social, barandiaran2008adaptivity, moreno2018minimal, levin2019computational}. This motivation is at the center of the functioning of individuals and since early childhood drives goal-oriented behavior \cite{csibra2008goal,sommerville2005pulling}. The achievement of goals is fundamental for individuals \cite{abele2007agency,abele2014communal} and is related to outcomes such as social status, career success, and well-being \cite{deci2000and,stajkovic1998self,gebauer2013agency}. However, goal achievement rarely occurs in isolation, as individuals often depend on others to facilitate or hinder their progress. This interdependence explains why humans are highly attuned to agency-related cues, such as biological motion \cite{simion2008predisposition} or causality, intentionality \cite{frith2010social}, which help navigate social dynamics and optimize cooperation or competition. Beyond these individual and interpersonal factors, agency is deeply embedded within a sociocultural framework that supports or limits one's ability to act and achieve goals \cite{ahearn2001language}. By linking individual capacities, social interdependence, and cultural influences, agency serves as a unifying construct to understand collective dynamics. Consequently, agency plays a crucial role in fostering collective action, as it underpins the belief of individuals in their ability to make a meaningful impact toward shared goals \cite{van2016building}. Psychological research highlights that people are more likely to participate in collective action when they believe that their efforts can lead to tangible results, reflecting a sense of achievable goals and personal influence \cite{drury2009collective,thomas2009aligning, abbe2018community,van2008toward}. We will rely on the link between agency and collective action in the validation study presented below. 

Agency can be automatically measured from text samples using BERTAgent (\href{https://pypi.org/project/bertagent}{pypi.org/project/bertagent}), a recent \ac{NLP} \Tom{tool based on the \ac{LLM}} RoBERTa \cite{liu2019roberta}, trained on a large labeled dataset of samples generated from existing agency dictionaries available from the literature, and properly validated by a panel of experts \cite{nikadon2023}. BERTAgent has been shown to outperform existing approaches for the automated identification of agency levels in text, and is therefore taken as our reference pre-trained \Tom{\acp{LLM} to test algorithms} capabilities. It comes as a ready-to-use Python package. It has recently been used to predict political participation in elections \cite{nikadon2024}, call-to-action behavior before and after a significant event \cite{erseghe2023projection}, and depression states through its negative connotation \cite{witkowska2025diminished}.

\Section[MM]{Methods and materials}

We first review the algorithms \Tom{that were selected and tested for explainability at the word level, together with the chosen quality measures available from the literature that were used to compare performance and identify appropriate system parameters. A brief description of the datasets used is also given}.

\subsection{Word-level attribution methods for explainability}

\Tom{An accurate review of state-of-the-art methods available from the literature, and concerning the explainability of \acp{LLM} classifiers and regressors at the word level, let us identify a set of techniques that include the standard \ac{IG} approach. Here, we briefly review the basics of these approaches, as they will be used in what follows to test their suitability in the chosen scenario.}
Specifically, we consider the following.
\begin{itemize}
\item {\bf \ac{IG}}: This is the reference \ac{IG} algorithm of \cite{sundararajan2017axiomatic}, implementing gradient integration along a straight path from a baseline input $\B x_0$ to the tested input $\B x$ in the form
$$
\B a(\B x)  = \int_{\Bs x_0}^{\Bs x} \nabla F(\B u)\,d\B u\;,
\e{IG2}
$$
where $\B a$ is the \emph{attribution} output, $F$ is the classification model (\Tom{e.g., based on RoBERTa or any other state-of-the-art \ac{LLM}}), and $\nabla$ is the gradient operator. $\B x_0$ and $\B x$ represent the input embeddings of words in a sentence. The variable $\B u$ is an intermediate input vector along the straight-line path connecting the baseline $\B x_0$ to the tested input $\B x$. \ac{IG} possesses the \emph{completeness} property
$$
\sum_i a_i(\B x) = F(\B x)-F(\B x_0)\;.
\e{IG4}
$$
This ensures that the attribution sums equal the output of the transformer model whenever the baseline output is zero, $F(\B x_0)=0$, which is a desired property. \ac{IG} is implemented via successive evaluations of the gradient along the line connecting $\B x_0$ to $\B x$, in the form
$$
\B a(\B x) \simeq \B\Delta \times\sum_{k=0}^{N} \nabla F(\B x_0+k\B\Delta ) \;,
\e{IG6}
$$
where $\B\Delta =\frac1N(\B x-\B x_0)$ is the step vector, $N$ is the number of steps in the Riemann approximation of the integral, $\times$ is an entry-wise multiplication among vectors,  and $\nabla$ is the gradient numerically computed by backpropagation.
\item {\bf Sequential \ac{IG}}: This method is based on the \ac{IG} approach, but calculates the importance of each word in a sentence by creating an ad hoc baseline $\B x_0$ for that word, where every other word is held fixed and the reference word is replaced by a MASK token embedding \cite{enguehard2023sequential}. Due to its iterative nature, it adds complexity to the standard \ac{IG} approach and also loses the completeness property.
\item {\bf Discretized \ac{IG}}: This approach uses a more effective attribution along nonlinear interpolation paths, based on the rationale that ``\emph{straight-line interpolated points are not representative of text data due to the inherent discreteness of the word embedding space}'' \cite{sanyal2021discretized}. Although it retains the completeness property, it is computationally expensive and requires searching for \emph{close} words in the embedding space. Moreover, it does not provide any consistent improvement over \ac{IG} when run on a RoBERTa model \cite{sanyal2021discretized}, and for this reason, it will not be considered further in this paper.
\item {\bf DeepLIFT}: This method extends the \ac{IG} concept by backpropagating activation differences, in such a way as to permit the information to propagate even when the gradient is zero, allowing us to identify dependencies missed by other methods \cite{shrikumar2017learning}.
\item {\bf GradientSHAP}: This technique exploits Shapley values, a concept from game theory, to attribute an importance score to each part of the input \cite{mosca2022shap, scott2017unified,park2024stress}. It has been applied to different approaches ranging from \ac{IG} (GradientSHAP) to DeepLIFT (DeepSHAP). In this paper, we consider the SHAP implementation enhancing \ac{IG}.
\end{itemize} 
All \Tom{the attribution methods} are tested in Python using the open source PyTorch CAPTUM library (\href{https://captum.ai}{captum.ai}) and run on GPU resources.

As previously discussed, \ac{IG} based approaches are dependent on two main parameters, the number of steps $N$ and the baseline input $\B x_0$. We therefore test them for optimal performance by varying these parameters according to the following approach:
\begin{itemize}
\item {\bf Number of steps}: the number of integration step is known to have an impact on the quality of \ac{IG} attributions, with the correct value largely dependent on the specific application \cite{makino-etal-2024-impact}; a reference range $N\in[20,1000]$ is available for applications of the Transformer architecture\Tom{, which is the reference \ac{LLM} architecture}, and we therefore take it as our reference for testing purposes.

\item {\bf Baseline input}: The choice of the baseline input should ensure that $F(\B x_0)$ is as close as possible to the zero value, to allow the sum of the association values to correspond to the target output $F(\B x)$. For this reason, the baseline embedding $\B x_0$ is built to retain the CLS, SEP, and PAD token embeddings (so that their association values remain zero) while setting the remaining embeddings to a reference value chosen among the following inspired by the literature (e.g., see \cite{bastings2021will}):
\begin{itemize}
\item {\bf zero}: the baseline embedding is filled with zero values.
\item {\bf mask}: the baseline embedding is filled with MASK token embeddings.
\item {\bf padding}: the baseline embedding is filled with PAD token embeddings.
\item {\bf mean}: the baseline embedding is filled with the mean value calculated on the whole set of RoBERTa input embeddings.
\end{itemize}
The baseline choice is known to have an impact on the performance, and only an appropriate baseline choice is able to generate reliable attributions.
\end{itemize}

\subsection{Quality measures for test purposes}

For comparison purposes between different methods and different settings, we selected \Tom{from the state-of-the-art solutions available in the literature} two faithfulness measures from \cite{deyoung2020benchmark}, namely \emph{comprehensiveness} and \emph{sufficiency}, as well as an \emph{approximation error} that evaluates the deviation from the output of the ideal model \cite{makino-etal-2024-impact}. Faithfulness metrics evaluate whether the important features identified by the algorithms are genuinely important (comprehensiveness) and sufficient (sufficiency) to make predictions. These metrics are assessed by masking features in the order of their importance scores and analyzing how the model predictions change \cite{benchmark24eai}. \Tom{Overall, the chosen metrics cover the relevant aspects of faithfulness, and are widely adopted and recognized in the literature as reliable and effective.} We have:
\begin{itemize}
\item {\bf Comprehensiveness}: The output is tested against $\B x$ and a counterpart $\B x_f^c$ where a fraction $f$ of words is removed. The words removed correspond to the \emph{top} scoring attribution values (in absolute value) \cite{yin2022sensitivity}. We calculate
$$
C_f(\B x) = | F(\B x)-F(\B x_f^c)|\;.
\e{IG10}
$$
Comprehensiveness naturally increases with $f$, and a higher comprehensiveness indicates better algorithm performance, as the best scores contribute more significantly to the prediction.
\item {\bf Sufficiency}: The output is tested against $\B x$ and a counterpart $\B x_f^s$, which is the complement of $\B x_f^c$. In $\B x_f^s$, the fraction $f$ of top-scoring attribution values (in absolute value) is retained, while the remaining words are removed \cite{yin2022sensitivity}. We calculate
$$
S_f(\B x) = |F(\B x)- F(\B x_f^s)|\;.
\e{IG12}
$$
Sufficiency naturally decreases with $f$, and lower sufficiency values indicate better algorithm performance.
\item {\bf Approximation error}: This is a normalized measure \cite{makino-etal-2024-impact}
$$
AE = \left| \frac{ \sum_i a_i(\B x) - (F(\B x)-F(\B x_0))}{F(\B x)-F(\B x_0)}\right| \;,
\e{IG14}
$$
which measures the deviation of the actual computed value from the theoretical value. This metric equals zero in the case of ideal integration and captures the deviation from ideality based on the choice of the number of steps $N$.
\end{itemize}

\subsection{Datasets}

In order to test different \Tom{attribution methods}, we selected a few datasets covering distinct content, as well as distinct sentence ranges. Specifically, we have:
\begin{itemize}
\item {\bf \#covid19} a dataset of COVID-19 Twitter posts collected from February 11 to April 11, 2020, using the hashtag \#covid19 in the search; 1000 tweets per week were selected among the most influential, i.e. those that received the most reactions, replies, likes and quotes; the dataset is taken from \cite{erseghe2023projection}.
\item {\bf IMDB} a set of 50K reviews of highly popular movies, allowing no more than 30 reviews per movie, typically used for binary sentiment classification \cite{maas2011learning};  the dataset was collected in 2011, and contains an even number of positive and negative reviews; it is available from HuggingFace.
\item {\bf slurs} a set of 336 real or fictitious stories written by female participants in a 2022 questionnaire, describing a situation in which they experienced verbal aggression; the data is a subset including the two experimental conditions described in \cite{cervone2025words}.
\item {\bf snippets} a collection of very short Twitter posts of a maximum of 5 words taken from the TweetEval dataset; specifically, we use ``SemEval 2018 Task 1: Affect in Tweets'' dataset, which focuses on detecting emotional content in social media posts \cite{mohammad2018semeval}.
\end{itemize} 
\begin{figure}[t]
\centerline{\includegraphics[height=5.1cm, angle=0]{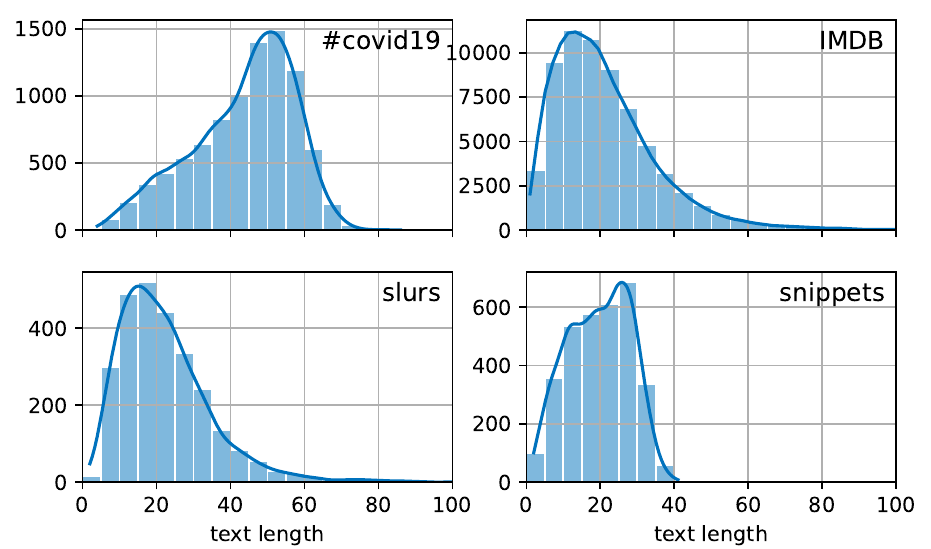}}
\caption{Number of tweets/sentences as a function of their length, in the four datasets.}
\label{fig:IG-IG2}
\end{figure}
To ensure correct input to our regressors or classifiers, the datasets ``IMDB'' and ``slurs'', which contain very long texts, were split into sentences to control the text length. A superficial cleaning was also applied to all datasets in order to remove mentions, HTML links, special characters, and double spaces, where present. The text length distribution (number of tokens distribution) of the datasets is reported in \fig{IG2}. Note how ``IMDB'' and ``slurs'' show an equivalent distribution (at the sentence level) mostly active in the range from 5 to 40, while the distribution of the ``\#covid19'' dataset, where we kept the original tweets since they have a strict length limit, is active in a higher range, from 30 to 70. For the ``snippets'' database, the distribution is active in the lowest range of 0 to 40.

\begin{figure*}[t]
\centerline{\includegraphics[height=9cm, angle=0]{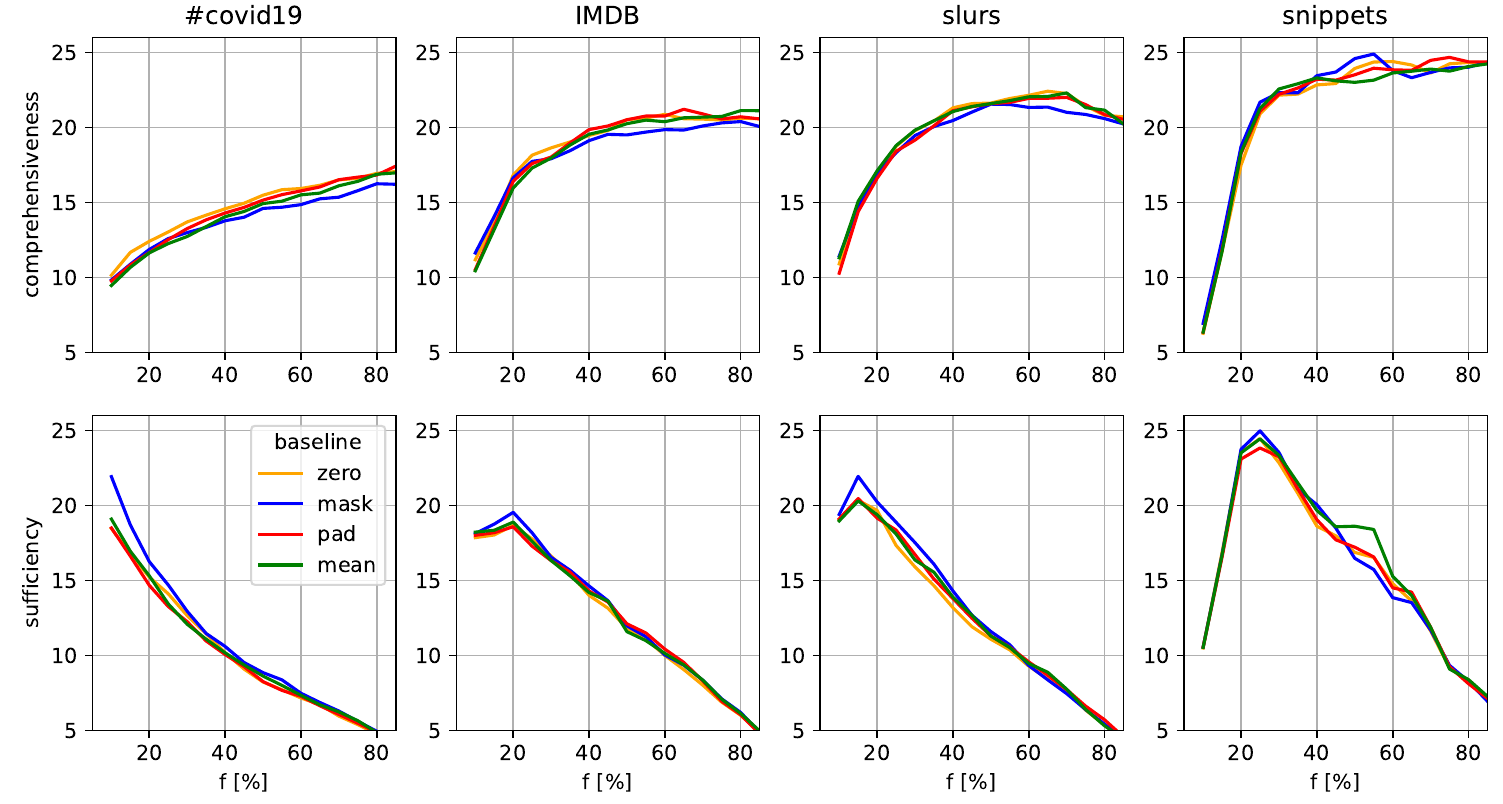}}
\caption{Comprehensiveness and sufficiency in \ac{IG} (with $N=300$ steps) as a function of the fraction $f$ in the four datasets, and for different baseline choices.}
\label{fig:IG-IG4}
\end{figure*}

To further control the computational burden of the test campaign, we randomly selected a subset of text samples from each dataset to capture the entire range of active text lengths. Specifically, $1600$ tweets equally distributed in the length range $[25,64]$ were selected from the ``\#covid19'' dataset ($40$ samples per text length), $1500$ sentences equally distributed in the length range $[1,49]$ from the ``IMDB'' dataset ($30$ samples per text length), $1335$ sentences equally distributed in the length range $[5,40]$ from the ``slurs'' dataset (approximately $40$ samples per text length), and, finally, $300$ tweets equally distributed in the length range $[2,14]$ from the ``snippets'' dataset (approximately $20$ samples per text length). In this way, the snippets dataset targets short text samples, while the other datasets cover a wider range.

\Tom{For verification purposes, we selected a dataset from the development of BERTAgent \cite{nikadon2023}, namely:
\begin{itemize}
\item {\bf golden standard dictionary}, a collection of texts written by participants in experimental settings aimed to induce high agency or low agency language production. We observe that this collection was used for verification also in the development of BERTAgent, hence it is completely uncorrelated to the LLM we are building on.
\end{itemize}}

An additional dataset was used to verify the potential of the explainability approach, namely:
\begin{itemize}
\item {\bf highlights} a collection of 802 short texts (100–125 words) written by native English speakers (authors) on a socially relevant topic they deemed personally important. Authors were asked to ``\emph{review the list of socially relevant issues below \footnote{List of socially relevant issues: \emph{raise awareness of mental health},  \emph{prevent climate change and protect environment}, \emph{increase voting turnout in elections}, \emph{reduce economic inequality}, \emph{increase volunteering}, \emph{advocate for free speech}, \emph{protect human rights}, \emph{ensure food security and sustainable agriculture}, and \emph{advocate for digital privacy and security}.} and select the one that resonates most with you and you believe it requires mobilizing others to take action in order to address it}.'' 
After completing the writing task, authors highlighted the text fragments they considered most effective in mobilizing others. An identical highlight task was assigned to 2,347 independent participants (readers), with at least two readers evaluating each text (M$=2.4$, SD$=0.76$) \cite{Nikadonetal2025}. The data used in the present paper were collected before December 3\textsuperscript{rd}, 2024.
\end{itemize}

\Tom{We incidentally observe that, in the paper, the datasets ``\#covid19,'' ``IMDB,'' and ``snippets'' are only used as a (rich) data source for comparing the performance of different attribution methods, and for selection purposes, but we are not offering new insights on them. Instead, the ``slurs'' and ``highlights'' datasets are exploited in the application examples of, respectively, \sect{AO} and \sect{TO} where they are analyzed under the novel lens discussed in this paper.}

\Section[IG]{Attribution method selection}

\Tom{In this section we provide a detailed performance comparison between the attribution methods identified in the literature, covering performance and complexity aspects, which will ultimately identify \ac{IG} as the preferred solution. The comparison is here targeted to the explainability of socio-psychological markers, and we take agency as our reference marker both because of its general importance, and because it is the only marker for which a validated tool (verified by expert psychologists) exists.}

\subsection{Parameters selection in IG}

\Tom{Before being able to compare different attribution methods, an appropriate selection of parameters must be determined for the \ac{IG} approach, which we do in the following.}
The performance of \ac{IG} is illustrated in \fig{IG4} and \fig{IG12} with respect to the measures of comprehensiveness and sufficiency, in \fig{IG6} with respect to the value of the baseline output and in \fig{IG10} with respect to the approximation error. We comment on them in detail in the following.

\begin{figure}[t]
\centerline{\includegraphics[height=7.5cm, angle=0]{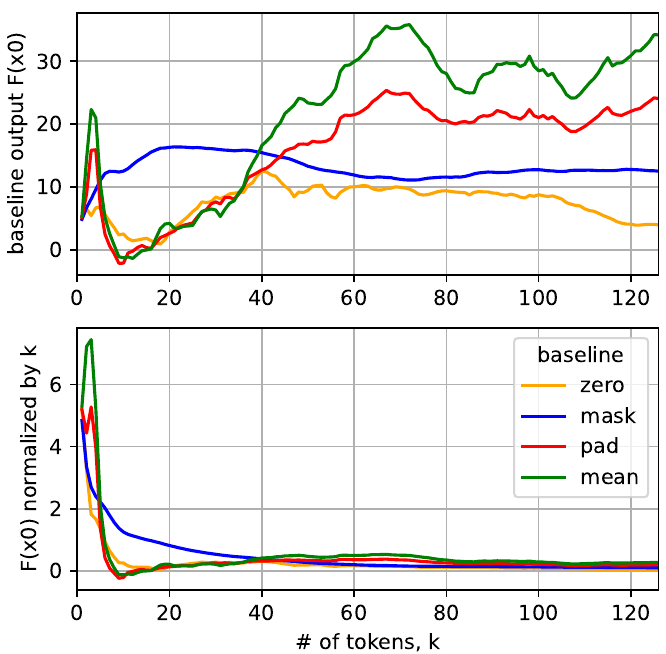}}
\caption{Baseline output $F(\B x_0)$ (above) and normalized baseline output (below) in \ac{IG} (with $N=300$ steps) as a function of the number $k$ of tokens in a text, and for different baseline choices.}
\label{fig:IG-IG6}
\end{figure}

\fig{IG4} shows comprehensiveness and sufficiency of the different datasets for the four different baseline choices. Here, the number of steps is fixed to $N=300$, to guarantee a reliable outcome. As is evident from \fig{IG4}, the performance dependency on the baseline choice is weak, since all behave very similarly. However, some interesting conclusions can be drawn. First, the \emph{mask} baseline (in blue in the figure) consistently shows the worst behavior in both comprehensiveness (lower curve) and sufficiency (higher curve, at least in the lower range of $f$). An analysis of the error regions (not shown in the figure for the sake of readability) further shows that the difference is statistically significant, that is, the \emph{mask} baseline regularly provides the worst performance. \emph{zero}, \emph{mean}, and \emph{padding} instead provide the best performance and essentially show equivalent behavior.   

To further investigate the choice of the baseline, \fig{IG6} shows the value of the baseline output $F(\B x_0)$ as a function of the length of the text (number of tokens $k$). In this case, the lower the baseline output value $F(\B x_0)$, the better the approach, since the deviation of the sum of attributions with respect to the true value $F(\B x)$ is controlled (see also \e{IG4}). Above in \fig{IG6} we display the true value of the baseline output, which is then normalized by the text length $k$ below in the figure in order to capture the average contribution on each single token, that is, its real impact. From the figure, we clearly see that the \emph{mask} baseline (in blue) shows pretty constant behavior that, however, is constantly outperformed by the \emph{zero} baseline (in yellow). \emph{Padding} baseline (in red) and \emph{mean} baseline (in green) exhibit a similar behavior, with a peak at low $k$, and a high saturation value at large $k$, with \emph{mean} providing the highest values. They are also outperformed by the \emph{zero} baseline, except for a very small region in the interval $k\in[10,20]$ where, however, we observe the lowest values of $F(\B x_0)$. For this reason, the \emph{zero} baseline is the preferred choice, as it exhibits the best or the highest performance throughout $k$.

Some further insights are provided in \fig{IG12} and \fig{IG10} regarding the choice of the number of steps $N$. In general, as we can appreciate from \fig{IG12}, the role of the parameter $N$ is not captured by comprehensiveness and sufficiency measures, at least for $N$ in a reasonable range $N\ge50$. In fact, the curves in \fig{IG12} almost perfectly superpose. We observe that, although in \fig{IG12} we show a result for the ``snippets'' dataset with zero baseline only, this behavior was found to be consistent over the different datasets and the different baseline choices, although we do not show this explicitly, as it carries little information.

\begin{figure}[t]
\centerline{\includegraphics[height=5.0cm, angle=0]{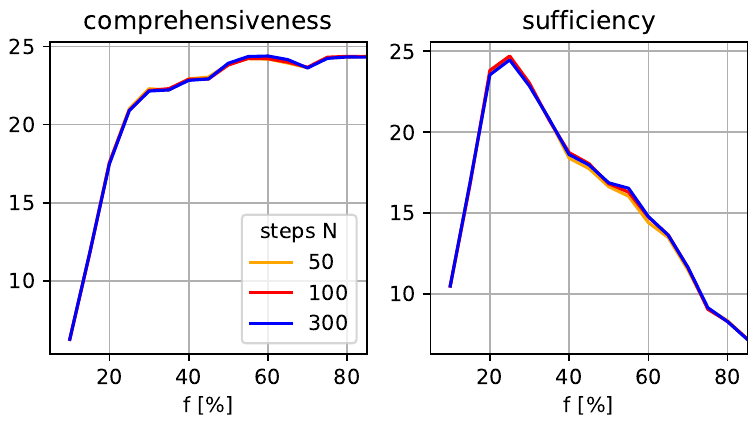}}
\caption{Comprehensiveness and sufficiency in \ac{IG} as a function of the fraction $f$ in the ``snippets'' dataset with zero baseline, and for different number of steps $N$.}
\label{fig:IG-IG12}
\end{figure}

\begin{figure*}[p]
\begin{center}
\bf \#covid19 + IMDB + slurs + snippets datasets\\
\includegraphics[height=5.0cm, angle=0]{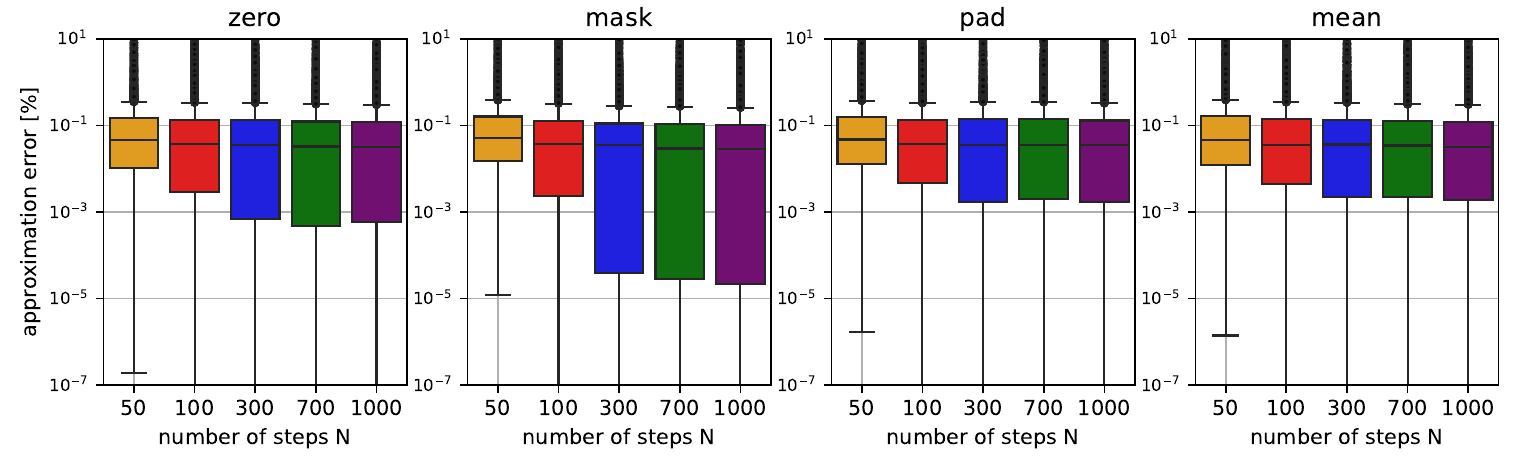}\\
\bf snippets dataset\\
\includegraphics[height=5.0cm, angle=0]{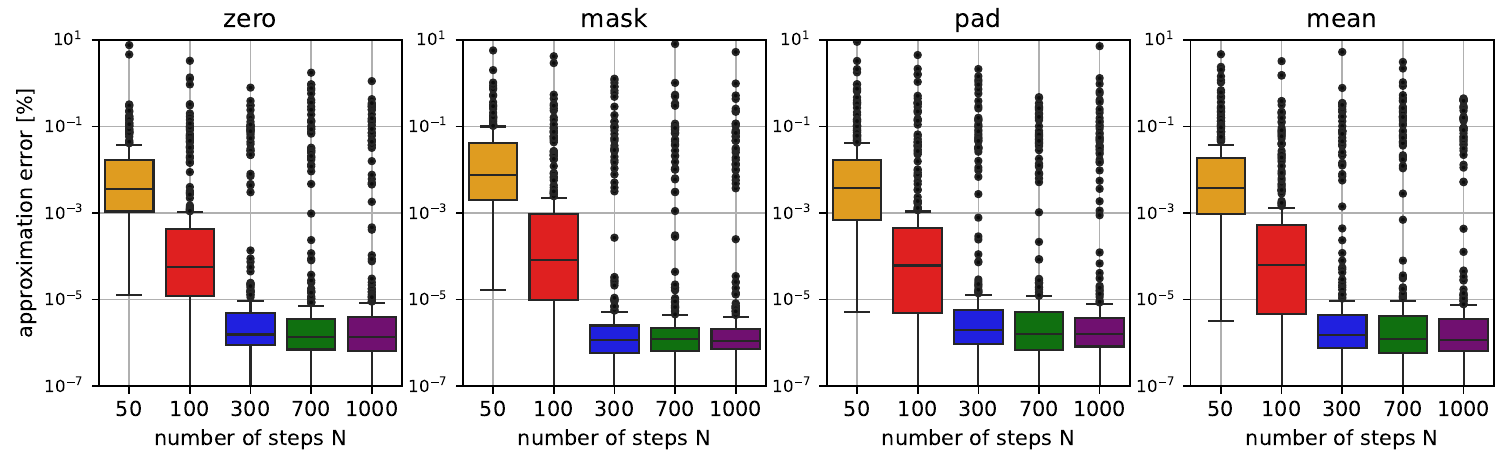}
\end{center}
\caption{Approximation error statistics in \ac{IG} as a function of the number of steps $N$ in the four datasets, and for different baseline choices.}
\label{fig:IG-IG10}
\end{figure*}

\begin{figure*}[p]
\centerline{\includegraphics[height=9cm, angle=0]{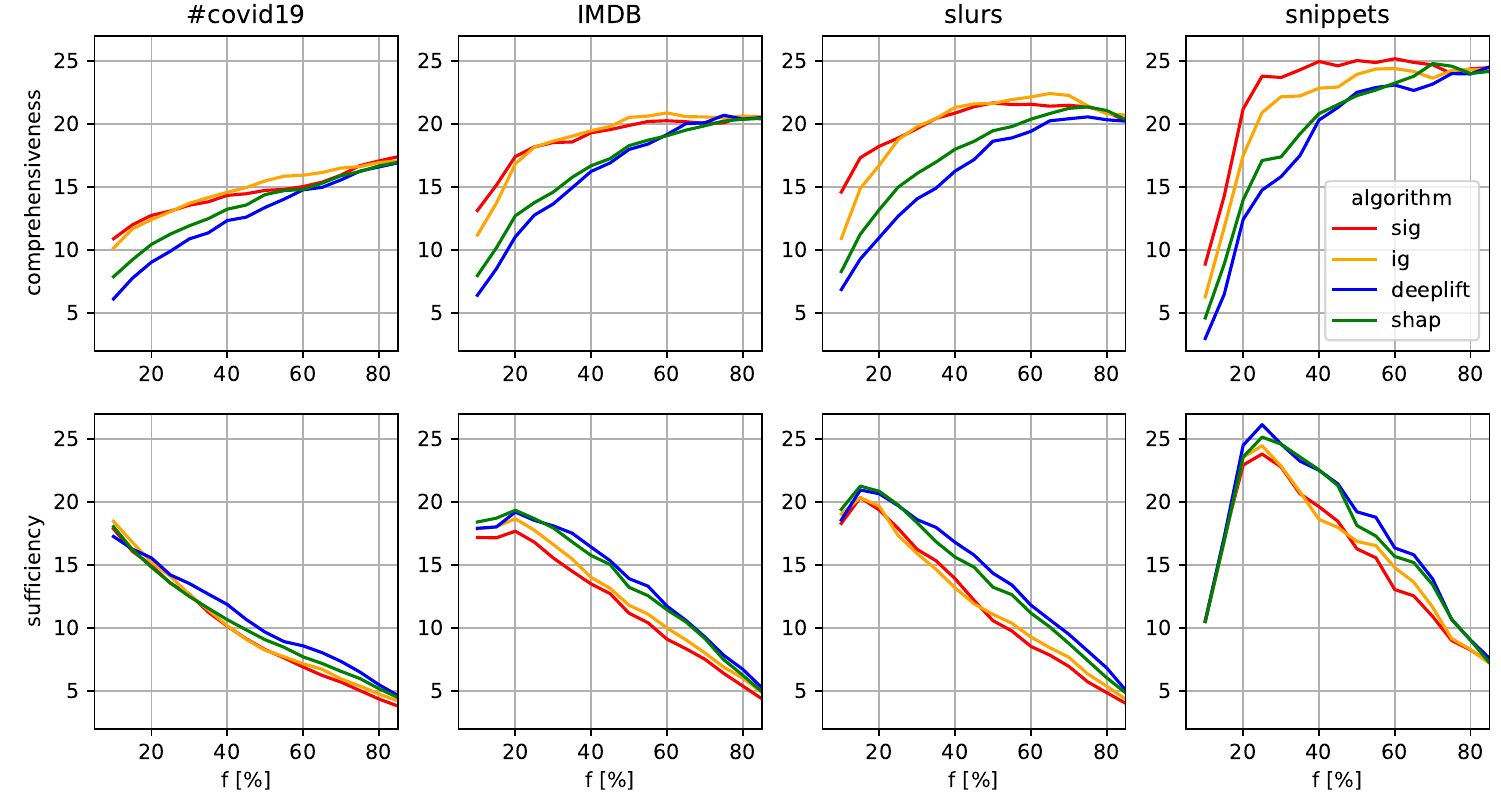}}
\caption{Comprehensiveness and sufficiency in different approaches as a function of the fraction $f$ in the four datasets.}
\label{fig:IG-IG33}
\end{figure*}

The role of the number of steps $N$ is better captured by the approximation error statistics displayed in \fig{IG10} for $N$ that spans $50$, $100$, $300$, $700$, and $1000$, and for different choices of the baseline. We separately report the statistic for the aggregated datasets (above) and the Snippets dataset only (below). The box plot reports the 25\%, 50\%, and 75\% quartiles (colored box) and outliers. From the plot above in \fig{IG10} we clearly see that the best performance is achieved with \emph{mask}, a small loss is experienced by the \emph{zero} baseline, while the \emph{padding} and \emph{mean} baselines provide the worst performance. Note how the performance tends to saturate for a number of steps greater than or equal to $N=300$, especially for \emph{zero} and \emph{mask}. This saturation effect, which we verified to be associated with the performance for short sentences, is evidenced by the plot below in \fig{IG10}, practically setting $N=300$ as a reliable choice.

In conclusion, the most reliable choice for \ac{IG} is identified in the \emph{zero} baseline, as it consistently provides among the best performance in comprehensiveness and sufficiency (see \fig{IG4}), the most reliable baseline output values (see \fig{IG6}) and among the best performance in approximation error (\fig{IG10}). The number of steps is set to $N=300$, as this choice provides a highly reliable outcome, especially for short sentences (see \fig{IG10}, below).

\subsection{Attribution methods comparison}

The comparison of \ac{IG} with alternative approaches is reported in \fig{IG33} and \fig{IG33a} for \emph{SequentialIG} (sig in figure), \emph{DeepLIFT}, and \emph{GradientSHAP} (shap in figure). In the figures, \ac{IG} is implemented with a \emph{zero} baseline and $N=300$, \emph{SequentialIG} with a \emph{zero} baseline and $N=300$ for the snippets dataset and $N=50$ for the other datasets due to its high computational complexity, while \emph{DeepLIFT} and \emph{GradientSHAP} are run with their reference implementation. 

\begin{figure}[t]
\centerline{\includegraphics[height=5.0cm, angle=0]{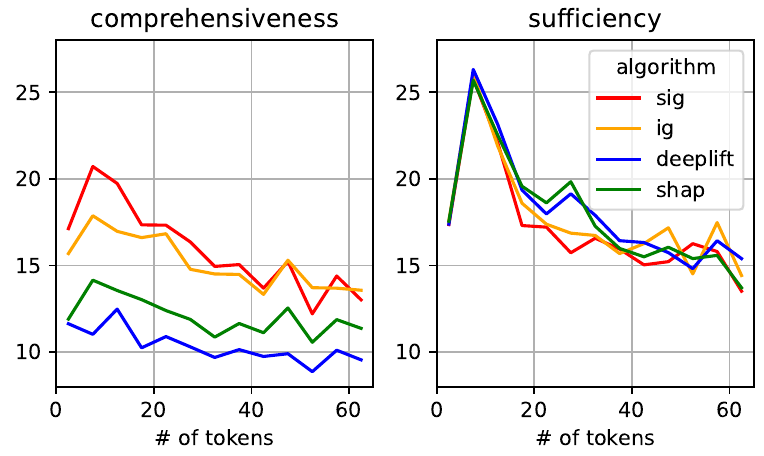}}
\caption{Comprehensiveness and sufficiency for $f=20\%$ in different approaches as a function of the number of tokens; all four datasets where used.}
\label{fig:IG-IG33a}
\end{figure}

In the comparison given in \fig{IG33}, we can clearly observe how \emph{DeepLIFT} and \emph{GradientSHAP} consistently provide the worst performance in terms of both comprehensiveness and sufficiency, in the four data sets, which is consistent with the findings of \cite{sanyal2021discretized}. In the comparison between \ac{IG} and \emph{SequentialIG}, instead, the expected improvement of the sequential approach does not occur, with the exception of comprehensiveness in the ``snippets'' dataset. A further inspection of the result as a function of the number of tokens, available in \fig{IG33a}, confirms this trend: an improvement is available only for comprehensiveness when considering very short text snippets of up to $10$ tokens in length. 

In general, the processing burden of \emph{DeepLIFT} and \emph{GradientSHAP} is about 10\% of that of \ac{IG} (they are very fast algorithms), while \emph{SequentialIG} runs at least 10 times longer than \ac{IG} due to its iterative nature. \Tom{This is illustrated in some detail in \tab{TIG3} where processing times are shown for the short sentences of the ``snippets'' dataset (but perfectly equivalent results apply to any dataset). Evidently, this outcome does not justify the computational burden of \emph{SequentialIG}, and identifies \ac{IG} as the preferred choice providing acceptable complexity and enhanced performance.}

\begin{table}[h]
\caption{\Tom{Average runtime (in seconds per sample) for each attribution method on the ``snippets'' dataset.}}
\label{tab:IG-TIG3}
\begin{center}\Tom{
\begin{tabular}{c|c}
\bf Attribution method & \bf Time (s)\\\hline
SequentialIG & 4.6552 \\
IG & 0.4542 \\
DeepLIFT & 0.0456\\
GradientSHAP & 0.0467 \\
\end{tabular}}
\end{center}
\end{table}

\Section[VE]{Validation step}

Having assessed \ac{IG} as \Tom{the reference approach according to standard state-of-the-art performance metrics}, some additional steps are needed to \Tom{properly validate its adequacy. To this aim, the raw \ac{IG} output is first processed for readability, then its effectiveness is evaluated by psychology experts. This allows us, on the one hand, to evaluate its overall quality and, on the other hand, to clearly identify its practical limits. We explain the procedure in the following.}

\subsection{Rendering the IG output to a readable form}
\label{rendering}

\Tom{We preliminarily address the need to graphically render the raw \ac{IG} output in a satisfactory visual form, which will then be used for fair validation. Specifically, some technical issues need to be solved, which are mainly linked to the specific way BERTAgent is implemented, and which we propose to approach as follows.}

The RoBERTa tokenizer often splits words into a number of tokens, for example 
$$
\hbox{\colorbox[HTML]{f7f6f7}{These} \colorbox[HTML]{f9eff4}{people} \colorbox[HTML]{faeaf2}{are} \colorbox[HTML]{aa0e68}{\color{white}lazy} \colorbox[HTML]{fce5f1}{and} \colorbox[HTML]{eeabd2}{unm} \colorbox[HTML]{fad6ea}{ot} \colorbox[HTML]{faeaf2}{ivated},}
\e{SP2}
$$
where darker highlights correspond to a stronger association value and where in this case it is the word ``unmotivated'' to be split into three tokens; to prevent this effect, we exploit the more readable token structure provided by SpaCy, and align the RoBERTa tokens to it by taking advantage of the offset information; specifically, every time a word is split into many tokens the corresponding attributions are added to provide a unique value, to get SpaCy associations of the form
$$
\hbox{\colorbox[HTML]{f7f6f7}{These} \colorbox[HTML]{f9eff4}{people} \colorbox[HTML]{faeaf2}{are} \colorbox[HTML]{aa0e68}{\color{white}lazy} \colorbox[HTML]{fce5f1}{and} \colorbox[HTML]{c51d7e}{\color{white}unmotivated}.}
\e{SP4}
$$

Some of the associations might have a sign which is not coherent with the overall agency value; for example, in a sentence that carries, overall, a positive agency, namely
$$
\hbox{\colorbox[HTML]{f7f7f6}{This} \colorbox[HTML]{f9f0f5}{person} \colorbox[HTML]{f5f7f2}{is} \colorbox[HTML]{3f811e}{\color{white}far} \colorbox[HTML]{6dad36}{\color{white}from} \colorbox[HTML]{e9f5d6}{being} \colorbox[HTML]{e07eb3}{lazy} \colorbox[HTML]{f1f6e8}{!}}
\e{SP8}
$$
the words ``person'' and ``lazy'' carry negative association values, but this might be confusing; for this reason, in the visual representation of agency at the word-level, we set to zero association all those values that carry a weight which is not coherent with the overall agency, to have
$$
\hbox{\colorbox[HTML]{f7f7f6}{This} \colorbox{white}{person} \colorbox[HTML]{f5f7f2}{is} \colorbox[HTML]{3f811e}{\color{white}far} \colorbox[HTML]{6dad36}{\color{white}from} \colorbox[HTML]{e9f5d6}{being} \colorbox{white}{lazy} \colorbox[HTML]{f1f6e8}{!}}
\e{SP10}
$$

Negations often carry most of the association weight although, conceptually, the association should also be linked to the word that is negated; for example, for the negative agency sentence
$$
\hbox{
\colorbox[HTML]{f6f7f5}{These} \colorbox[HTML]{eef6e2}{people} \colorbox[HTML]{f6f7f5}{are} \colorbox[HTML]{40831e}{\color{white}not} \colorbox[HTML]{ecf6de}{lazy} \colorbox[HTML]{f8f5f6}{at} \colorbox[HTML]{d9f0bc}{all}\colorbox[HTML]{ddf1c1}{.}
}
\e{SP6}
$$
the word ``not'' has most of the association value while, in practice, we would like the couple ``not lazy'' \Tom{or the triplet ``are not lazy''} to carry the positive agentic meaning. \Tom{In other cases, instead, only one keyword among a meaningful group is given a high association value. To solve this (interpretability) issue,} again we exploit SpaCy \Tom{dependencies to group, in the syntactic network, all those leaf nodes that share a common parent (by adding their agencies) in such a way to build meaningful syntactic groups that, for example, fully solve the negation problem. So, for the example given above we would obtain} 
$$
\hbox{\colorbox[HTML]{f7f7f6}{These} \colorbox[HTML]{f7f7f6}{people} \colorbox[HTML]{40831e}{\color{white}are not lazy} \colorbox[HTML]{f7f7f6}{at} \colorbox[HTML]{f7f7f6}{all}\colorbox[HTML]{f7f7f6}{.}
}
\e{SP6a}
$$

Finally, in order to facilitate the interpretation of the outcome, the maximum association value is normalized to the actual agency value $F(\B x)$, so that the lower agency is rendered by a weaker color, \Tom{punctuation is not given any visual association value, and all associations below the average association value are visually discarded (this improves readability on longer sentences)}; for example
$$
\hbox{\colorbox[HTML]{f7f7f6}{Join} \colorbox[HTML]{9ccf64}{our next \#fridays4future strike} \colorbox[HTML]{f7f7f6}{!}
}
\e{SP12}
$$
which, in the comparison with \e{SP10}, is rendered in a weaker green colour since it carries a lower agency level.

In general, the above preprocessing steps can hide the more technical aspects of \ac{IG} (and BERTAgent), while maintaining its core message. They allow a reliable visual interpretation of the association values and are, for this reason, suitable for properly interpreting the quality of the \ac{IG} approach. The reference code implementing these features is available at \url{https://github.com/aghababaeiali/sociomark-xai}.

\subsection{Validation procedure}

\Tom{Three expert psychologists were asked to evaluate the IG output (using optimized parameters:  \emph{zero} baseline and $N=300$ steps) with respect to:
\begin{itemize}
\item {\bf hits (HIT)}, that is parts of text correctly identified as carrying an agentic meaning, using a four point Likert scale ({\em none, some, many, all}) with the purpose of evaluating the correct detection capabilities of the approach;
\item {\bf false alarms (FA)}, that is parts of text incorrectly identified as carrying an agentic meaning, using a four point Likert scale ({\em none, some, many, all}) with the purpose of identifying hallucinations of the model;
\item {\bf exclusion flag}, that is a flag whose purpose is to detect, and then exclude, those texts where BERTagent has failed the sign of agency and are therefore not eligible to build a reliable statistic on attributions.
\end{itemize}
The joint information of HIT and FA constitutes a sufficiently rich information to allow a reliable validation analysis.}

\Tom{A selection of 200 texts from BERTAgent's ``golden standard dictionary'' was used, by selecting those text that carried a sufficiently high level of agency (absolute value above $0.2$), resulting in a sufficiently even distribution over the (absolute) range $0.2-0.8$. Out of the initial $200$ texts, $22$ samples were discarded by the exclusion flag ($13$ signaled by one expert, $8$ by two experts, and $1$ by all experts), so the final statistics rely on $n=178$ samples. The intercoder reliability was high between experts B and C (Cohen-K $0.63$(FA) $0.66$(HIT)), and weaker between experts A and B (Cohen-K $0.29$(FA) $0.21$(HIT)) and experts A and C (Cohen-K $0.30$(FA) $0.27$(HIT)), which is nevertheless expected given the complexity of the task requiring to effectively locate the sources of agency in the text.}

\subsection{Result}

\Tom{\fig{IG14} shows the behavior of the average score value (between coders) and its variance, highlighting a very high quality in HIT where the vast majority of texts are associated with higher levels with very limited standard deviation, and a more challenging conclusion with FA (hallucinations) that occur in 20\% of the text samples. This latter is a result perfectly coherent with the complexity of the task, which we ought to better understand.}

\begin{figure}[t]
\centerline{\includegraphics[height=6.5cm, angle=0]{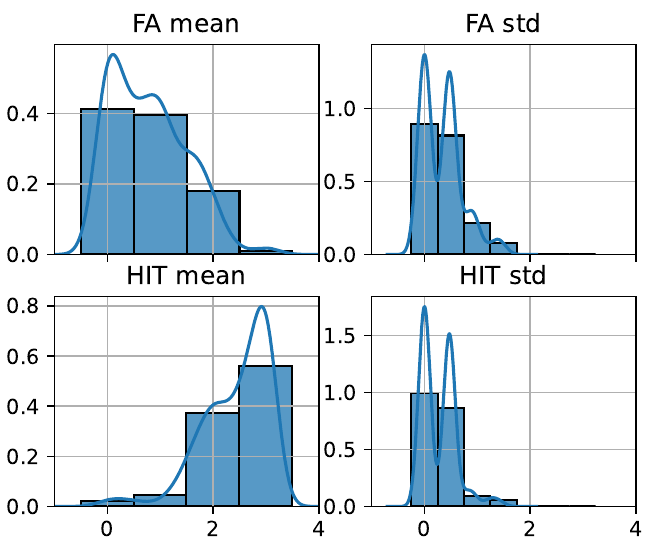}}
\caption{Average agreement between coders in terms of mean and standard deviation for HIT and FA, where the Likert scale has been mapped into numbers: 0-none, 1-some, 2-many, 3-all.}
\label{fig:IG-IG14}
\end{figure}

\Tom{To this aim, some interesting insights can be drawn from \fig{IG16}, showing the metrics dependence on the level of agency, separately for positive and negative agencies. As we can appreciate from the figure, the higher the agency level, the more reliable the algorithm, with performance slowly dropping as agency decreases (all intersect and slope factors were found to be statistically relevant here). Furthermore, an equivalent effect, but based on the text length, is reported in \fig{IG17}. Overall, the attribution method is therefore strongly reliable for larger values of agency (in the absolute sense) and shorter text lengths, while providing a performance degradation as the text length increases and the agency level decreases. These, however are not truly limiting factors, since we already observed a natural disagreement even between expert human coders in such a challenging context.}

\begin{figure}[t]
\centerline{\includegraphics[height=6.8cm, angle=0]{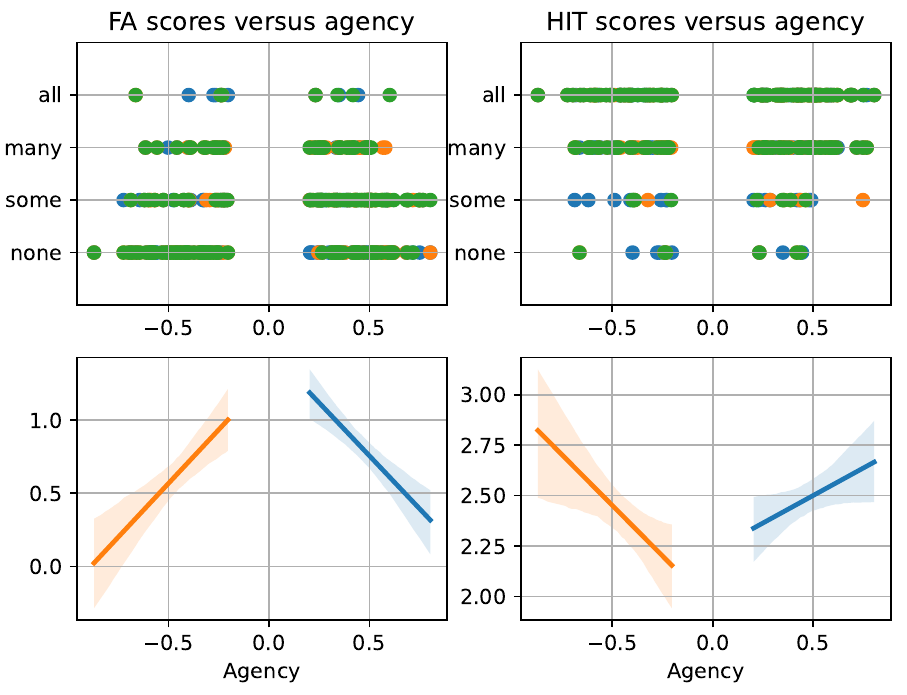}}
\caption{Dependency of FA and HIT scores on the overall agency value (above) and its linear fitting result (below) where shaded areas refer to a $95\%$ confidence interval.}
\label{fig:IG-IG16}
\end{figure}

\begin{figure}[t]
\centerline{\includegraphics[height=6.8cm, angle=0]{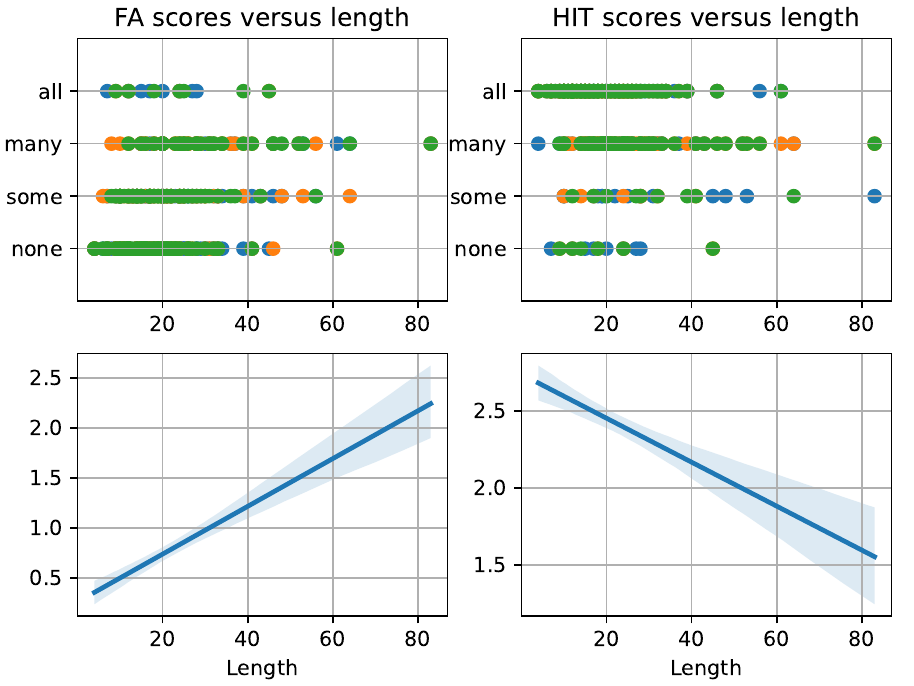}}
\caption{Dependency of FA and HIT scores on the text length (above) and its linear fitting result (below) where shaded areas refer to a $95\%$ confidence interval.}
\label{fig:IG-IG17}
\end{figure}

\Tom{Now, in order to further prove the effectiveness and plausibility of the chosen \ac{IG} approach, we finally discuss two meaningful application examples.}

\Section[AO]{Application \#1: Text highlights}

\Tom{In this first application example, we} exploit the \emph{highlights} dataset, namely a collection of 802 texts written by English native speakers on a specific subject (see Sect.~3.3 for dataset description) with the intention of motivating the reader to action, that is, naturally carrying a high level of agency. For each text, the author and two or three additional expert readers were asked to highlight which portions of the text carry the most relevant part of the call-to-action. \Tom{From socio-psychological theory, we} would therefore expect that the highlight captures the highest agency levels (at the word level), \Tom{which we can now measure through the attribution properties of \ac{IG} and the fact that they were verified to correctly capture places where agency occurs}.  

\begin{figure*}[t]
\centerline{\includegraphics[height=9.7cm, angle=0]{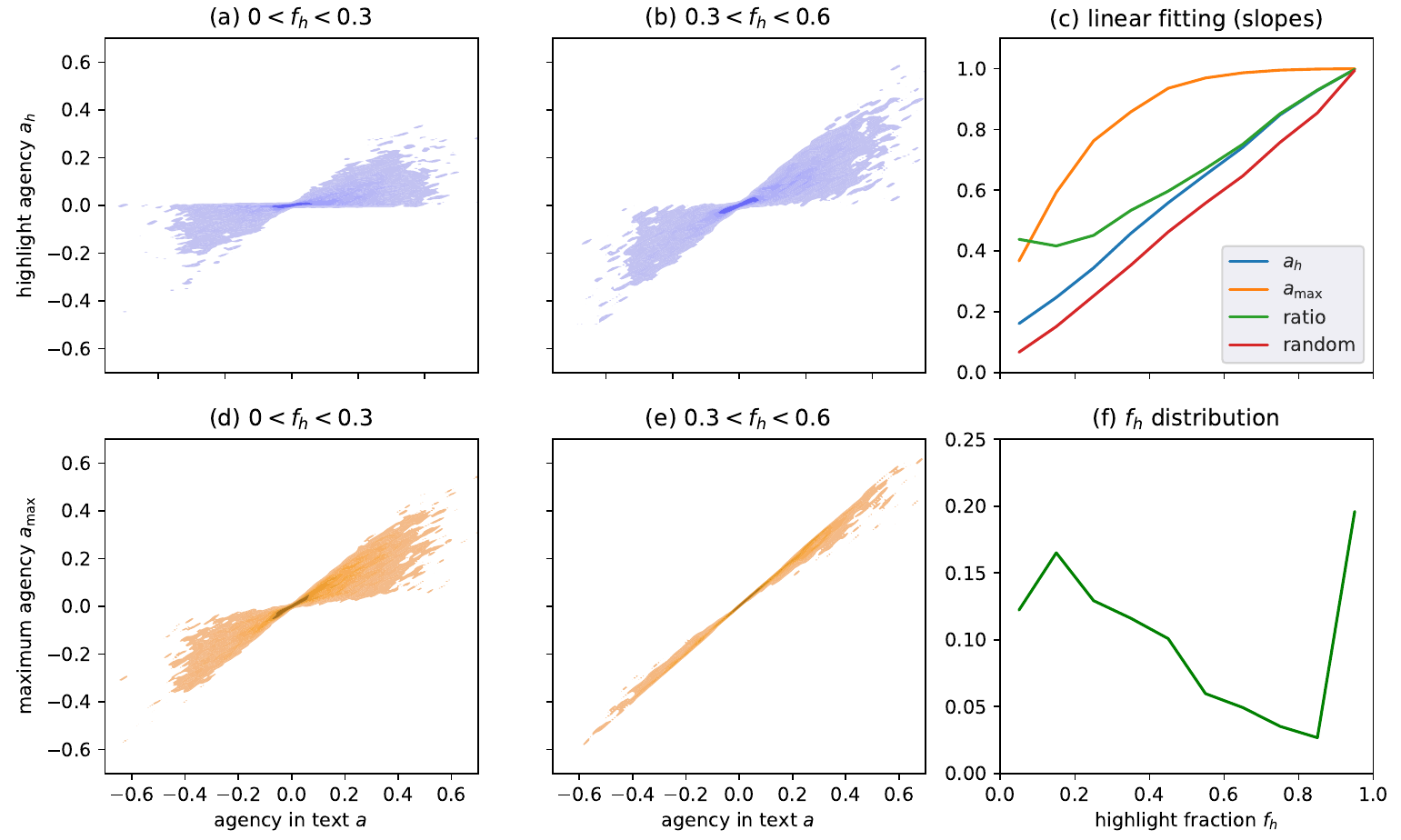}}
\caption{Highlights data: (a-b) highlight agency $a_h$ versus agency in text $a$ for $f_h\in(0,.3]$ and $f_h\in(.3,.6]$; (c) linear fitting (slopes) for highlight agency $a_h$ and maximum agency $a_{\max}$ versus agency in text $a$ as a function of the highlight fraction $f_h$; (d-e) maximum agency $a_{\max}$ versus agency in text $a$ for $f_h\in(0,.3]$ and $f_h\in(.3,.6]$; (f) distribution of the highlight fraction $f_h$.}
\label{fig:IG-IG20}
\end{figure*}

To our aim, the texts were divided into sentences and only those sentences that were (partially or totally) highlighted were retained. For each sentence, the following measures were identified:
\begin{itemize}
\item the overall agency level, $a$;
\item the highlight fraction $f_h$, namely what fraction of words were highlighted, ranging from $0$ to $1$;
\item the amount of agency contained in the highlight, $a_h$, evaluated via the \ac{IG} attribution output; for the sake of readability of the result, the attribution output was preliminarily polished according to the rendering procedure described in Section~\ref{rendering}, were, in particular, in a positive (negative) agentic text all negative (positive) attributions were set to zero and the resulting values were rescaled in order for their sum to provide the true agency level $a$;
\item the maximum agency level $a_{\max}$ for the considered highlight, that is, the sum of the strongest agency levels up to the highlight length, with a polishing procedure equivalent to the one applied for $a_h$;
\item the agency level captured in case as many words as in the highlight are selected, but these are selected at random in such a way to capture the overall noise level.
\end{itemize}
Observe that we have $a_h\le a_{\max}\le a$ for a positive agentic text and reverse ordering for a negative agentic text.

The values of the highlight agency $a_h$ and the maximum agency $a_{\max}$ are shown versus the overall agency $a$ on the left of \fig{IG20} for two different ranges of the highlight fraction $f_h$. Note that in all the considered cases the behavior is naturally captured by a linear fitting (slope only), the slope being obviously higher for the maximum agency $a_{\max}$ \Tom{(orange plots, below)} as it captures all the best choices. A more detailed overview of this idea is shown in the graph of \fig{IG20}.(c) where slope values of the linear fitting of $a_h$ and $a_{\max}$ versus $a$
consider a highlight fraction range of the form $f_h\in(\frac{k}{10},\frac{k+1}{10}]$. The plot also shows, in green, the ratio between the slopes of $a_h$ and $a_{\max}$, revealing that the highlight captures a consistent fraction of the most agentic words in the text, which is higher than $40\%$ even for very low values of the highlight fraction $f_h$. Moreover, the fact that $a_h$ is clearly separate from the noise level (with noise accounting for about $40\%$ of $a_h$ at low $f_h$)  certifies that the highlights naturally evidence the agentic portion of the text. We note incidentally that the highlight process, especially at low values of $f_h$, naturally implies a selection of the part of text that is both more agentic and more meaningful, and therefore it is natural not to necessarily capture the entire value $a_{\max}$. As a clarifying example, for the sentence
$$
\eqalign{
& \hbox{\colorbox[HTML]{f7f7f6}{If} \colorbox[HTML]{f7f7f6}{we} \colorbox[HTML]{f7f7f6}{all} \colorbox[HTML]{5ea02c}{\color{white}work together} \colorbox[HTML]{f2f6ec}{,}} \cr
&\hbox{ \colorbox[HTML]{93c959}{\color{black}\underline{we can make}} \colorbox[HTML]{f7f7f6}{\underline{a}} \colorbox[HTML]{f7f7f6}{\underline{difference}} \colorbox[HTML]{f7f7f6}{\underline{!}}
}}
\e{SP20}
$$
where the highlight (underlined text) is the final part, the expert reader is evidently selecting the agentic part of the text that is most meaningful to them. In this respect, an agreement with $a_{\max}$ greater than $40\%$ is a strong certification that the highlight process is invaluably linked to high agency in the text (as we expect from the theory), as well as that the \ac{IG} approach is fully capable of correctly capturing it.

This finding is significant for sociopsychological research. It demonstrates that individuals, when considering collective mobilization, spontaneously highlight agency as a critical factor in creating an effective mobilization message. This observation is consistent with questionnaire-based studies \cite{van2008toward} showing that the efficacy (agency) consistently predicts collective action intentions ($r=.34$; it accounts for $11.56\%$ of the variance in collective action intentions). In this study, we are further emphasizing its role in motivating collective action. Recent evidence \cite{formanowicz2023mobilize} also underscores the importance of agentic language in mobilization strategies, linking it to the effectiveness of such messages. The prevalence of mobilizing speech acts fosters a shared understanding between communicators and audiences, consistent with the persuasion knowledge model \cite{friestad1994persuasion}. This shared knowledge enhances the use and recognition of linguistic features that characterize persuasive attempts, thereby contributing to their effectiveness. The presented method provides insight into the specific elements that drive this effectiveness, elucidating how communication fosters collective action, engagement, and compliance.

\Section[TO]{Application \#2: Dictionaries from labels}

As we have largely discussed so far, the \ac{IG} approach is able to correctly highlight the keywords that drive the output values in a classification or regression \ac{NLP} task. We proved it in detail with agency and the BERTAgent regressor, but the idea is evidently applicable over a wide range (for example, for sentiment or any other sociopsychological marker of interest). However, the drawback of what we have considered so far is the availability of a large labeled dataset for training our \ac{NLP} algorithm. In many applications, this large dataset is not available and we are instead constrained by a small dataset that can be easily labeled in practice. 

In this latter scenario of small labeled datasets, the \ac{IG} technique can be exploited to reveal the keywords related to the different classes identified by the labels, or, in case the label is a continuous value, to reveal the keywords that concur in building different levels of such value. \Tom{This approach, for example, would be able to build (or expand) dictionaries on specific targets (e.g., sexualization, irony, etc) which are in general complex to deal with. We explain this application example in detail in the following.}

\subsection{Rationale for small labeled datasets}

We set ourselves a task of \emph{explainability}, which we formalize as follows. 
\begin{itemize}
\item First, we train a \Tom{\ac{LLM} (e.g., RoBERTa)} to detect labels (or values in case we build a regressor), 
\item then we exploit \ac{IG} to identify the keywords related to each label (or regressor level), to obtain a thorough description of the classes and, more importantly, capturing those keywords that differentiate one class from the others. 
\end{itemize}
In order for this approach to work correctly, we need to address the training process in a somewhat unusual way. Unlike what we have learned from the \ac{ML} literature, here we not only allow but deliberately \emph{encourage overfitting}, up to a point where the trained \ac{NLP} model exhibits full precision (accuracy of $1$). In this context, there is also no need to differentiate between a train and a test set, as the intention is that of overfitting; hence, all the available data can be used as a train set. In this way, we are confident that the model we are training fully captures the statistics that are peculiar to each class and specific to the dataset. \Tom{Note that this procedure is rather common in setups where deep learning \ac{NLP} models are used as optimization frameworks, e.g., in the context of topic analysis \cite{joo2020dirichlet,gui2020multi}, where the approach was proven to be robust.}

Specifically, the \ac{NLP} classification task here is performed by a ``RobertaForSequenceClassification'' model in PyTorch, trained to accuracy $1$ (in about $20$ epochs) with GPU processing resources. The \ac{NLP} classifier is fed the raw text. \ac{IG} is then run on the trained model to extract word-level associations that carry a description of each class. 
In the following, we demonstrate that this peculiar solution works effectively using a standard application to a dataset labeled for relevant sociopsychological features.

\subsection{Application scenario}

The idea is here applied to the ``slurs'' dataset which, as explained, contains a collection of $336$ real or fictitious stories describing a situation of verbal aggression and harassment suffered by a female subject. Sexual objectification is a phenomenon by which people are reduced to their body, sexual body parts, or sexual functions \cite{fredrickson1997obj}. Women are frequent targets of sexual objectification, with estimates suggesting that they are treated as sexual objects as frequently as 1.3 times per day \cite{holland2017sexual}, with important consequences on their life outcomes and self-perceptions: for one, repeated exposures to sexual objectification can lead to the internalization of this external perspective on the self (i.e., self-objectification).  Both objectification and self-objectification have severe consequences on women's wellbeing \cite{koval2019obj}, including a higher risk of health issues such as depression and eating disorders \cite{roberts2018objectification}. 

The dataset was labeled by social psychology experts for relevant aspects, namely the gender of the aggressor (male, female, both) and the relationship between the victim and the aggressor (acquaintance, friend, partner/family, unknown). Furthermore, it includes participants' ratings (in levels from 1 to 7) concerning the emotions they experienced during the event (e.g., shame and fear), as well as the extent to which they felt treated as a sexual object. 

\begin{table*}[p!]
\caption{Slurs dataset: Keywords identified by \ac{IG} under different labels.}
\label{tab:IG-TIG2}
\input{TIG2}
\end{table*}

\subsection{Dictionaries from sociopsychological labels}

The application of the proposed method to the above labels is illustrated in \tab{TIG2}, where only the top 20 scoring words are shown for each class, with a darker color indicating a stronger association value. Several unlabeled texts are also present and have been included in the classification process, although their IG values are not shown in the tables because they carry little meaning. 

For the first label, \emph{gender of the aggressor}, we appreciate from the top of \tab{TIG2} that the proposed technique works correctly, being able to efficiently identify all keywords that are markers of gender. As such, for aggressions conducted by a male perpetrator, male-related keywords such as \emph{male}, \emph{boy}, \emph{brother}, \emph{boyfriend}, \emph{man}, etc. are identified. In the woman class, female-related keywords are also correctly associated with verbs such as \emph{obsess} and \emph{charge}, which are needed to distinguish female keywords referring to victims (always women) from keywords referring to the perpetrator. The second label, \emph{relationship between the victim and the aggressor}, further corroborates the accuracy of the tool. We see in \tab{TIG2} that the partner/family class correctly identifies \emph{daughter}, \emph{partner}, \emph{boyfriend}, \emph{husband}, \emph{dad}, \emph{sister}, etc., in the keyword list. Also, a correct distinction is available between the acquaintance class, including figures related to the subject such as \emph{employer}, \emph{boss}, \emph{mate}, \emph{advisor}, \emph{supervisor}, and the unknown person class which instead includes more generic words like \emph{salesperson}, \emph{customer}, \emph{gentleman}, etc.

In the realm of sexual objectification, \emph{shame} and \emph{fear} are two key emotional outcomes \cite{szymanski2021interpersonal}. In line with this previous literature, we see in \tab{TIG2} that the high-level classes correctly identify words related to shame (e.g., \emph{mortify}, \emph{insecurity}) and fear (e.g., \emph{scare}, \emph{afraid}), both of which are not present in the low-level classes. This is important because it attests to the sensitivity of IG to emotion gradients. Accordingly, high levels of \emph{fear} are denoted by references to aggression (e.g., \emph{violence}, \emph{unprovoked}, \emph{abusive}). The \emph{objectification} label, at the bottom of \tab{TIG2}, also provides compelling information. In particular, high levels of objectification are featured by sexual content (e.g., \emph{blowjob}, \emph{cock}, \emph{slut}) and clothing (\emph{skirt}), in line with the fact that sexual objectification is characterized by the reduction of a person to their sexual body parts or functions \cite{gervais2012seeing}, which translates to a focus on physical appearance above one's personhood.

Overall, not only does the proposed technique display the criteria that one would expect to guide human evaluations, and thus proves to be a valuable tool, but it also provides insights about the most relevant keywords in classifications of complex phenomena, such as objectification. Although the classification of male vs. female characters in a story could also be achieved with a rather simple \ac{NLP} tool, and emotions can be detected through dictionaries, a complex sociopsychological phenomenon such as self-objectification entails a more nuanced appraisal of the relevant keywords. In particular, within the objectification process, the observer's perspective on one's body is blended with the exploitation of the dehumanized person as a sexual toolkit. This technique could therefore also be applied to build or expand new dictionaries, including words that would possibly not emerge as synonyms of the initial keywords but semantically related as specific clues of the construct under investigation.

\Section[CO]{Conclusion}

In this paper we approached the problem of explainability at the word-level for relevant sociopsychological semantic markers, that include but are not limited to the common sentiment analysis. By a thorough comparison of \Tom{state-of-the-art solutions and an accurate choice of their parameters}, we identified \acf{IG} as the most reliable choice, and developed \Tom{and validated} an \ac{IG}-based tool for the word-level identification of agency, a fundamental marker identifying collective dynamics that is particularly relevant, e.g., in social and political discourses. 

We further \Tom{discussed the application of} the \ac{IG} machinery to a more focused dataset scenario, where labels can be manually \Tom{coded}, and the \ac{IG} explainability can be applied to identify words that semantically characterize each class. Although, in general, the quality of the output strongly depends on the specific label and the availability of strong (and statistically coherent) semantic content that can be captured by the \ac{NLP} training procedure, the technique is relevant in at least two ways. First, for all those classification tasks lacking a strong and mature theoretical ground (e.g., the objectification task discussed in the paper, but also fake news detection, vaccine hesitancy behavior and causes, polarization mechanics in online social networks, etc.) can benefit in understanding which semantic content characterizes the classes, in such a way as to be able to extend existing dictionaries and strengthen or widen the theoretical ground. Second, more pragmatically, the semantic content highlighted by \ac{IG} can be used to extend, by similarity, a dataset in a recursive fashion until coherent detection is achieved. But these aspects are left to our future work.






\bibliographystyle{IEEEbib}
\bibliography{ig}





\vfill

\end{document}

%% file: TIG2.tex
\begin{center}
\begin{tabular}{|r|c|p{0.65\textwidth}|}
\hline
\bf class & \bf \#docs & \bf keywords\\\hline\hline

\multicolumn{3} {|c|}
{{\bf GENDER} (of the aggressor) - classifier 4 classes}\rule{0mm}{5mm}\\[2mm]\hline\hline
both & 5 & \textcolor[HTML]{276419}{couple} \textcolor[HTML]{c4e699}{husband} \textcolor[HTML]{cfebaa}{walk} \textcolor[HTML]{dbf0bf}{dog} \textcolor[HTML]{e1f3c7}{people} \textcolor[HTML]{e7f5d2}{block} \textcolor[HTML]{e7f5d3}{aggressive} \textcolor[HTML]{e7f5d3}{woman} \textcolor[HTML]{e8f5d5}{front} \textcolor[HTML]{e9f5d8}{ask} \textcolor[HTML]{e9f5d8}{voice} \textcolor[HTML]{eaf5d9}{person} \textcolor[HTML]{ebf6dc}{think} \textcolor[HTML]{ebf6dc}{partner} \textcolor[HTML]{ecf6de}{hate} \textcolor[HTML]{edf6df}{house} \textcolor[HTML]{edf6e1}{come} \textcolor[HTML]{eef6e2}{yell} \textcolor[HTML]{eef6e2}{friend} \textcolor[HTML]{eff6e4}{stand}\\
man & 189 & \textcolor[HTML]{276419}{male} \textcolor[HTML]{276419}{boy} \textcolor[HTML]{3d7f1e}{brother} \textcolor[HTML]{40831e}{boyfriend} \textcolor[HTML]{43861f}{man} \textcolor[HTML]{4e9322}{lad} \textcolor[HTML]{569927}{guy} \textcolor[HTML]{5c9e2a}{gentleman} \textcolor[HTML]{62a32e}{husband} \textcolor[HTML]{64a52f}{dad} \textcolor[HTML]{69aa33}{notice} \textcolor[HTML]{77b53c}{target} \textcolor[HTML]{81bd44}{accord} \textcolor[HTML]{83bf46}{annoying} \textcolor[HTML]{83bf46}{misogynistic} \textcolor[HTML]{86c049}{boss} \textcolor[HTML]{8fc654}{sudden} \textcolor[HTML]{9ed067}{supervisor} \textcolor[HTML]{a1d26a}{entirely} \textcolor[HTML]{a1d26a}{kid} \\
woman & 92 & \textcolor[HTML]{276419}{obsess} \textcolor[HTML]{276419}{secretary} \textcolor[HTML]{276419}{waitress} \textcolor[HTML]{286619}{sister} \textcolor[HTML]{43861f}{girl} \textcolor[HTML]{569927}{girlfriend} \textcolor[HTML]{569927}{nurse} \textcolor[HTML]{5c9e2a}{mother} \textcolor[HTML]{5ea02c}{woman} \textcolor[HTML]{64a52f}{charge} \textcolor[HTML]{69aa33}{supermarket} \textcolor[HTML]{71b038}{staff} \textcolor[HTML]{75b43b}{granddaughter} \textcolor[HTML]{7bb93e}{bedroom} \textcolor[HTML]{7dba40}{restaurant} \textcolor[HTML]{7dba40}{owner} \textcolor[HTML]{88c24c}{intervene} \textcolor[HTML]{8ac34f}{treat} \textcolor[HTML]{8fc654}{same}
\\
NA & 50 & \\
\hline\hline

\multicolumn{3} {|c|}
{{\bf RELATIONSHIP} (between the victim and the aggressor) - classifier 5 classes}\rule{0mm}{5mm}\\[2mm]\hline\hline
acquaintance & 62 & \textcolor[HTML]{276419}{employer} \textcolor[HTML]{31711b}{boss} \textcolor[HTML]{36761c}{mate} \textcolor[HTML]{3f811e}{advisor} \textcolor[HTML]{40831e}{thesis} \textcolor[HTML]{42841f}{supervisor} \textcolor[HTML]{67a832}{roommate} \textcolor[HTML]{6bac34}{master} \textcolor[HTML]{6faf37}{neighbour} \textcolor[HTML]{83bf46}{neighbor} \textcolor[HTML]{86c049}{colleague} \textcolor[HTML]{93c959}{university} \textcolor[HTML]{98cc5f}{classmate} \textcolor[HTML]{98cc5f}{single} \textcolor[HTML]{a3d36c}{use} \textcolor[HTML]{a7d672}{client} \textcolor[HTML]{a7d672}{catechism} \textcolor[HTML]{acd977}{exercise} \textcolor[HTML]{acd977}{factory} \textcolor[HTML]{aeda7a}{guy}\\
friend & 26 & \textcolor[HTML]{276419}{friend} \textcolor[HTML]{2b691a}{friendship} \textcolor[HTML]{6faf37}{boyfriend} \textcolor[HTML]{8fc654}{classmate} \textcolor[HTML]{95cb5c}{university} \textcolor[HTML]{98cc5f}{college} \textcolor[HTML]{9acd61}{good} \textcolor[HTML]{9ccf64}{classroom} \textcolor[HTML]{a5d56f}{mutual} \textcolor[HTML]{b5df82}{back} \textcolor[HTML]{b9e187}{believe} \textcolor[HTML]{bbe28a}{impulsively} \textcolor[HTML]{bee490}{m} \textcolor[HTML]{c0e593}{talk} \textcolor[HTML]{c0e593}{take} \textcolor[HTML]{cbe9a4}{call} \textcolor[HTML]{cdeaa7}{still} \textcolor[HTML]{cdeaa7}{due} \textcolor[HTML]{cdeaa7}{tell} \textcolor[HTML]{cdeaa7}{year}\\
partner/family & 62 & \textcolor[HTML]{276419}{daughter} \textcolor[HTML]{276419}{engage} \textcolor[HTML]{286619}{partner} \textcolor[HTML]{306f1b}{date} \textcolor[HTML]{36761c}{boyfriend} \textcolor[HTML]{37781c}{husband} \textcolor[HTML]{3f811e}{dad} \textcolor[HTML]{43861f}{sister} \textcolor[HTML]{5a9d29}{ex} \textcolor[HTML]{5c9e2a}{brother} \textcolor[HTML]{64a52f}{girlfriend} \textcolor[HTML]{71b038}{mother} \textcolor[HTML]{77b53c}{obsess} \textcolor[HTML]{7bb93e}{guy} \textcolor[HTML]{88c24c}{spend} \textcolor[HTML]{9acd61}{then} \textcolor[HTML]{9ccf64}{conversation} \textcolor[HTML]{9ccf64}{pregnant} \textcolor[HTML]{9ed067}{mum}\\
unknown person & 166 & \textcolor[HTML]{276419}{forum} \textcolor[HTML]{276419}{salesperson} \textcolor[HTML]{33721c}{checkout} \textcolor[HTML]{488c20}{store} \textcolor[HTML]{529624}{customer} \textcolor[HTML]{549825}{counter} \textcolor[HTML]{5ea02c}{gentleman} \textcolor[HTML]{67a832}{motorist} \textcolor[HTML]{69aa33}{aged} \textcolor[HTML]{71b038}{overtook} \textcolor[HTML]{75b43b}{pizza} \textcolor[HTML]{7bb93e}{winning} \textcolor[HTML]{7fbc41}{service} \textcolor[HTML]{83bf46}{catch} \textcolor[HTML]{86c049}{vaxxer} \textcolor[HTML]{8ac34f}{exchange} \textcolor[HTML]{8ac34f}{colleague} \textcolor[HTML]{8ac34f}{citizen} \textcolor[HTML]{8cc551}{bar} \textcolor[HTML]{95cb5c}{nearby}
\\
NA & 20 & \\\hline\hline

\multicolumn{3} {|c|}
{{\bf SHAME} (level of shame evaluated by authors) - classifier 7 classes}\rule{0mm}{5mm}\\[2mm]\hline\hline
low (1-2) & 124 & \textcolor[HTML]{276419}{confrontational} \textcolor[HTML]{276419}{incense} \textcolor[HTML]{276419}{overtook} \textcolor[HTML]{276419}{unprovoked} \textcolor[HTML]{4e9322}{excuse} \textcolor[HTML]{549825}{chap} \textcolor[HTML]{5a9d29}{offer} \textcolor[HTML]{62a32e}{prostitute} \textcolor[HTML]{64a52f}{scooter} \textcolor[HTML]{66a731}{chauvinist} \textcolor[HTML]{67a832}{mutual} \textcolor[HTML]{67a832}{mph} \textcolor[HTML]{69aa33}{junk} \textcolor[HTML]{69aa33}{derisive} \textcolor[HTML]{73b239}{ban} \textcolor[HTML]{73b239}{butcher} \textcolor[HTML]{7bb93e}{annoying} \textcolor[HTML]{7dba40}{polite} \textcolor[HTML]{81bd44}{impatient} \textcolor[HTML]{83bf46}{honk}\\
medium (3-5) & 112 & \textcolor[HTML]{276419}{cleavage} \textcolor[HTML]{40831e}{pushy} \textcolor[HTML]{43861f}{desk} \textcolor[HTML]{498d20}{oppose} \textcolor[HTML]{5ea02c}{sketchy} \textcolor[HTML]{69aa33}{embarrasment} \textcolor[HTML]{69aa33}{embarrsse} \textcolor[HTML]{6dad36}{hurtful} \textcolor[HTML]{71b038}{reach} \textcolor[HTML]{71b038}{reschedule} \textcolor[HTML]{73b239}{division} \textcolor[HTML]{75b43b}{effort} \textcolor[HTML]{77b53c}{bitchy} \textcolor[HTML]{77b53c}{pizza} \textcolor[HTML]{81bd44}{report} \textcolor[HTML]{83bf46}{particularly} \textcolor[HTML]{86c049}{tipsy} \textcolor[HTML]{8ac34f}{offensive} \textcolor[HTML]{8ac34f}{tonight} \textcolor[HTML]{8cc551}{wickedness}\\
high (6-7) & 100 & \textcolor[HTML]{276419}{assign} \textcolor[HTML]{276419}{heartbroken} \textcolor[HTML]{276419}{infatuation} \textcolor[HTML]{276419}{mortify} \textcolor[HTML]{276419}{speechless} \textcolor[HTML]{4c9121}{normal} \textcolor[HTML]{529624}{insecurity} \textcolor[HTML]{5ea02c}{inattentive} \textcolor[HTML]{62a32e}{dad} \textcolor[HTML]{64a52f}{university} \textcolor[HTML]{79b73d}{behaviour} \textcolor[HTML]{7bb93e}{girlfriend} \textcolor[HTML]{7dba40}{compulsive} \textcolor[HTML]{7fbc41}{syllabus} \textcolor[HTML]{7fbc41}{humiliate} \textcolor[HTML]{83bf46}{heavily} \textcolor[HTML]{8ac34f}{breathe} \textcolor[HTML]{8ac34f}{terrible} \textcolor[HTML]{8ac34f}{salesperson} \textcolor[HTML]{91c857}{fear}
\\\hline\hline

\multicolumn{3} {|c|}
{{\bf FEAR} (level of fear evaluated by authors) - classifier 7 classes}\rule{0mm}{5mm}\\[2mm]\hline\hline
low (1-2) & 108 & \textcolor[HTML]{276419}{topic} \textcolor[HTML]{5c9e2a}{report} \textcolor[HTML]{5ea02c}{rude} \textcolor[HTML]{62a32e}{assign} \textcolor[HTML]{67a832}{employer} \textcolor[HTML]{6dad36}{join} \textcolor[HTML]{73b239}{cheat} \textcolor[HTML]{73b239}{amusing} \textcolor[HTML]{79b73d}{inattentive} \textcolor[HTML]{7bb93e}{blowjob} \textcolor[HTML]{7dba40}{oratory} \textcolor[HTML]{81bd44}{division} \textcolor[HTML]{81bd44}{item} \textcolor[HTML]{83bf46}{offense} \textcolor[HTML]{8cc551}{sell} \textcolor[HTML]{8cc551}{team} \textcolor[HTML]{95cb5c}{skirt} \textcolor[HTML]{98cc5f}{awkward} \textcolor[HTML]{9acd61}{jest} \textcolor[HTML]{9acd61}{random} \\
medium (3-5) & 138 & \textcolor[HTML]{276419}{heartbroken} \textcolor[HTML]{276419}{temper} \textcolor[HTML]{276419}{unqualified} \textcolor[HTML]{2b691a}{dickhead} \textcolor[HTML]{31711b}{spend} \textcolor[HTML]{33721c}{bully} \textcolor[HTML]{397a1d}{exactly} \textcolor[HTML]{397a1d}{loose} \textcolor[HTML]{488c20}{bias} \textcolor[HTML]{488c20}{overtook} \textcolor[HTML]{509423}{reckless} \textcolor[HTML]{5a9d29}{threaten} \textcolor[HTML]{5a9d29}{witness} \textcolor[HTML]{5c9e2a}{misogynistic} \textcolor[HTML]{5ea02c}{useless} \textcolor[HTML]{6bac34}{nightclub} \textcolor[HTML]{6faf37}{insecure} \textcolor[HTML]{71b038}{entitle} \textcolor[HTML]{73b239}{outfit} \textcolor[HTML]{73b239}{space}\\
high (6-7) & 90 & \textcolor[HTML]{276419}{bouncer} \textcolor[HTML]{276419}{violence} \textcolor[HTML]{276419}{violent} \textcolor[HTML]{276419}{unprovoked} \textcolor[HTML]{498d20}{scare} \textcolor[HTML]{4c9121}{scary} \textcolor[HTML]{549825}{afraid} \textcolor[HTML]{60a22d}{pizza} \textcolor[HTML]{6bac34}{chase} \textcolor[HTML]{71b038}{nervousness} \textcolor[HTML]{71b038}{abusive} \textcolor[HTML]{73b239}{avoid} \textcolor[HTML]{75b43b}{ground} \textcolor[HTML]{79b73d}{youngish} \textcolor[HTML]{7bb93e}{aggressive} \textcolor[HTML]{7dba40}{horn} \textcolor[HTML]{7fbc41}{terrify} \textcolor[HTML]{7fbc41}{demeanour} \textcolor[HTML]{81bd44}{memorable} \textcolor[HTML]{88c24c}{nightclub} \\\hline\hline

\multicolumn{3} {|c|}
{{\bf OBJECTIFICATION} (level of objectification evaluated by experts) - classifier 7 classes}\rule{0mm}{5mm}\\[2mm]\hline\hline
low (1-2) & 124 & \textcolor[HTML]{276419}{foolish} \textcolor[HTML]{276419}{grumpy} \textcolor[HTML]{397a1d}{incompetent} \textcolor[HTML]{3d7f1e}{excuse} \textcolor[HTML]{4e9322}{attribute} \textcolor[HTML]{5a9d29}{slacker} \textcolor[HTML]{64a52f}{intelligent} \textcolor[HTML]{66a731}{ungrateful} \textcolor[HTML]{67a832}{error} \textcolor[HTML]{69aa33}{log} \textcolor[HTML]{6dad36}{handicapped} \textcolor[HTML]{75b43b}{slur} \textcolor[HTML]{77b53c}{retard} \textcolor[HTML]{77b53c}{misunderstand} \textcolor[HTML]{7dba40}{housekeep} \textcolor[HTML]{83bf46}{cockroach} \textcolor[HTML]{83bf46}{suitcase} \textcolor[HTML]{83bf46}{disgusting} \textcolor[HTML]{86c049}{omission} \textcolor[HTML]{86c049}{manoeuvre}\\
medium (3-5) & 92 & \textcolor[HTML]{276419}{buggy} \textcolor[HTML]{468a20}{confrontational} \textcolor[HTML]{5ea02c}{muzzle} \textcolor[HTML]{62a32e}{roundabout} \textcolor[HTML]{64a52f}{reprimand} \textcolor[HTML]{66a731}{telepass} \textcolor[HTML]{69aa33}{cocksucke} \textcolor[HTML]{7dba40}{whore} \textcolor[HTML]{7dba40}{fight} \textcolor[HTML]{7fbc41}{slut} \textcolor[HTML]{86c049}{savage} \textcolor[HTML]{8cc551}{spineless} \textcolor[HTML]{8fc654}{huff} \textcolor[HTML]{91c857}{foul} \textcolor[HTML]{93c959}{agitate} \textcolor[HTML]{98cc5f}{ex} \textcolor[HTML]{98cc5f}{threaten} \textcolor[HTML]{9acd61}{sexism} \textcolor[HTML]{9acd61}{wound} \textcolor[HTML]{9acd61}{idiotic}\\
high (6-7) & 120 & \textcolor[HTML]{276419}{blowjob} \textcolor[HTML]{276419}{cock} \textcolor[HTML]{276419}{cunt} \textcolor[HTML]{276419}{infatuation} \textcolor[HTML]{276419}{pushy} \textcolor[HTML]{276419}{slut} \textcolor[HTML]{2e6d1b}{skirt} \textcolor[HTML]{3f811e}{moron} \textcolor[HTML]{40831e}{whore} \textcolor[HTML]{42841f}{cheat} \textcolor[HTML]{468a20}{wet} \textcolor[HTML]{509423}{slag} \textcolor[HTML]{67a832}{humiliation} \textcolor[HTML]{6dad36}{nightclub} \textcolor[HTML]{6faf37}{sweatshirt} \textcolor[HTML]{71b038}{pizza} \textcolor[HTML]{71b038}{real} \textcolor[HTML]{79b73d}{heart} \textcolor[HTML]{7bb93e}{dick} \textcolor[HTML]{7dba40}{jealous}
\\\hline\hline

\multicolumn{3} {|c|}
{{\bf SELF OBJECTIFICATION} (level of self objectification evaluated by experts) - classifier 9 classes}\rule{0mm}{5mm}\\[2mm]\hline\hline
low (-6,-2) & 96 & \textcolor[HTML]{276419}{handicapped} \textcolor[HTML]{2b691a}{arm} \textcolor[HTML]{3c7d1d}{mood} \textcolor[HTML]{42841f}{offer} \textcolor[HTML]{468a20}{attribute} \textcolor[HTML]{4c9121}{overtook} \textcolor[HTML]{589b28}{infant} \textcolor[HTML]{5a9d29}{health} \textcolor[HTML]{6bac34}{drunk} \textcolor[HTML]{71b038}{walker} \textcolor[HTML]{79b73d}{mental} \textcolor[HTML]{81bd44}{sheep} \textcolor[HTML]{83bf46}{staff} \textcolor[HTML]{86c049}{intervene} \textcolor[HTML]{8cc551}{sibling} \textcolor[HTML]{91c857}{landlord} \textcolor[HTML]{95cb5c}{criticize} \textcolor[HTML]{95cb5c}{yesterday} \textcolor[HTML]{98cc5f}{bar} \textcolor[HTML]{a1d26a}{refuse} \\
medium (-2,2) & 147 & \textcolor[HTML]{276419}{cook} \textcolor[HTML]{276419}{dick} \textcolor[HTML]{276419}{elevator} \textcolor[HTML]{276419}{shit} \textcolor[HTML]{276419}{thick} \textcolor[HTML]{306f1b}{pizza} \textcolor[HTML]{3c7d1d}{compulsive} \textcolor[HTML]{468a20}{macho} \textcolor[HTML]{488c20}{forum} \textcolor[HTML]{589b28}{employer} \textcolor[HTML]{5a9d29}{tender} \textcolor[HTML]{5ea02c}{system} \textcolor[HTML]{64a52f}{unprovoked} \textcolor[HTML]{67a832}{ride} \textcolor[HTML]{6faf37}{challenge} \textcolor[HTML]{79b73d}{deny} \textcolor[HTML]{7dba40}{seat} \textcolor[HTML]{7dba40}{loose} \textcolor[HTML]{7dba40}{wheelchair} \textcolor[HTML]{7fbc41}{key}  \\
high (2,6) & 93 & \textcolor[HTML]{276419}{blowjob} \textcolor[HTML]{276419}{cock} \textcolor[HTML]{468a20}{small} \textcolor[HTML]{498d20}{hire} \textcolor[HTML]{64a52f}{pregnant} \textcolor[HTML]{69aa33}{tipsy} \textcolor[HTML]{6dad36}{salesperson} \textcolor[HTML]{71b038}{single} \textcolor[HTML]{7fbc41}{bed} \textcolor[HTML]{81bd44}{client} \textcolor[HTML]{81bd44}{hurl} \textcolor[HTML]{81bd44}{cleavage} \textcolor[HTML]{83bf46}{cockroach} \textcolor[HTML]{83bf46}{strength} \textcolor[HTML]{8ac34f}{overtake} \textcolor[HTML]{8cc551}{superior} \textcolor[HTML]{8cc551}{dog} \textcolor[HTML]{95cb5c}{sleep} \textcolor[HTML]{a3d36c}{shoe} \textcolor[HTML]{a9d874}{skirt}\\\hline\hline

\end{tabular}
\end{center}